\documentclass{article} 
\usepackage{iclr2025_conference,times}

\usepackage{amsmath,amsfonts,bm}

\def\eqref#1{equation~\ref{#1}}

\def\1{\bm{1}}

\DeclareMathAlphabet{\mathsfit}{\encodingdefault}{\sfdefault}{m}{sl}
\SetMathAlphabet{\mathsfit}{bold}{\encodingdefault}{\sfdefault}{bx}{n}

\usepackage{hyperref}
\usepackage{url}
\usepackage{graphicx}
\usepackage{amsmath}
\usepackage{amsthm}
\usepackage{amssymb}
\usepackage{mathtools}
\usepackage{enumitem}
\usepackage{wrapfig}
\usepackage{tcolorbox}
\usepackage{booktabs}
\usepackage{lipsum}
\usepackage{verbatim}
\usepackage{subcaption}

\theoremstyle{plain}

\title{Self-Distillation Enables Continual Learning}

\author{
Idan Shenfeld$^{1\,2}$\thanks{Correspondence to \texttt{idanshen@mit.edu}.} \quad Mehul Damani$^1$ \quad Jonas Hübotter$^3$ \quad Pulkit Agrawal$^{1\,2}$ \\
\;$^1$MIT 
\;$^2$Improbable AI Lab 
\;$^3$ETH Zurich \\
}

\iclrfinalcopy 
\begin{document}

\maketitle

\begin{abstract}
Continual learning, enabling models to acquire new skills and knowledge without degrading existing capabilities, remains a fundamental challenge for foundation models. While on-policy reinforcement learning can reduce forgetting, it requires explicit reward functions that are often unavailable. Learning from expert demonstrations, the primary alternative, is dominated by supervised fine-tuning (SFT), which is inherently off-policy. We introduce \textbf{Self-Distillation Fine-Tuning (SDFT)}, a simple method that enables on-policy learning directly from demonstrations. SDFT leverages in-context learning by using a demonstration-conditioned model as its own teacher, generating on-policy training signals that preserve prior capabilities while acquiring new skills. Across skill learning and knowledge acquisition tasks, SDFT consistently outperforms SFT, achieving higher new-task accuracy while substantially reducing catastrophic forgetting. In sequential learning experiments, SDFT enables a single model to accumulate multiple skills over time without performance regression, establishing on-policy distillation as a practical path to continual learning from demonstrations. Code and Datasets are available at \url{http://idanshenfeld.com/SDFT}.
\end{abstract}

\section{Introduction}
\label{introduction}

\begin{wrapfigure}{r}{0.5\textwidth}
    \vspace{-4ex}
    \centering
    \includegraphics[width=0.5\textwidth]{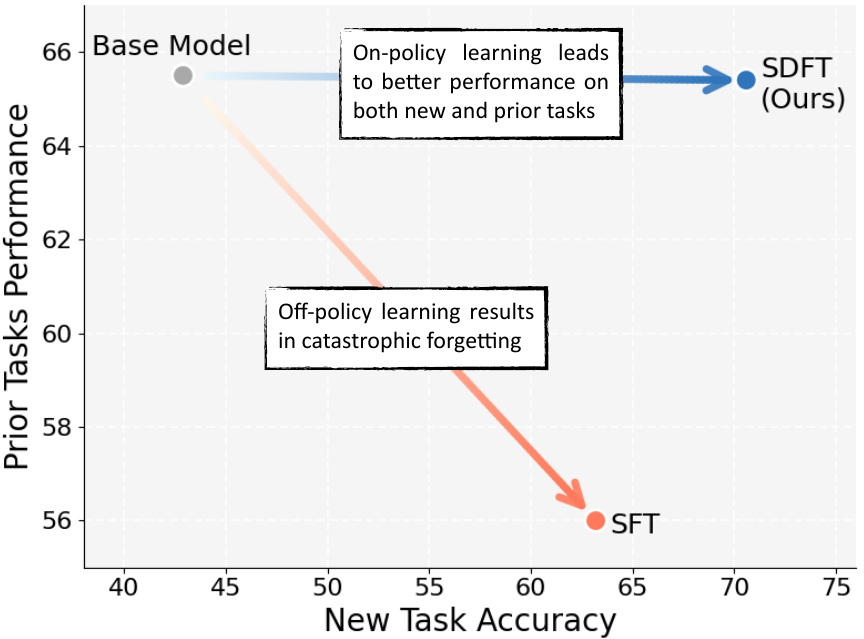}
    \vspace{-3ex}
    \caption{
    Supervised Fine-Tuning~(SFT) is commonly used to learn from expert demonstration datasets, but its off-policy nature leads to catastrophic forgetting of general capabilities.
    We introduce Self-Distillation Fine-Tuning (SDFT), which turns expert demonstrations into on-policy learning signals by using a demonstration-conditioned version of the model as its own teacher. In this way, SDFT enables true continual learning with the model improving on new tasks as they arise without regressing existing capabilities.}
    \label{fig:res_teaser}
    \vspace{-3ex}
\end{wrapfigure}

Foundation models have achieved remarkable success in recent years, powering AI applications across language, vision, robotics, and beyond. However, despite their impressive capabilities, today’s AI systems remain static after deployment. While they can adapt their behavior at inference time through mechanisms such as retrieval or prompting, they do not update their parameters to acquire new skills, internalize new knowledge, or improve from experience. To enable the next generation of foundation models, we must solve the problem of continual learning: enabling AI systems to keep learning and improving over time, similar to how humans accumulate knowledge and refine skills throughout their lives \citep{hassabis2017neuroscience, de2021continual}.

A growing body of recent work has highlighted the importance of on-policy learning for continual learning. When models learn from data generated by their current policy, they exhibit substantially reduced catastrophic forgetting compared to off-policy alternatives \citep{shenfeld2025rl, chen2025retaining}. To date, most successful on-policy approaches have been developed in the context of reinforcement learning (RL), where feedback is provided through an explicit reward function. However, in many real-world settings such rewards are unavailable or difficult to specify. Instead, learning typically proceeds from datasets of expert demonstrations. The dominant paradigm in this regime is supervised fine-tuning (SFT), which trains the model to imitate expert actions under a fixed, offline data distribution. While simple and scalable, SFT is inherently off-policy, and prior work has shown that sequential SFT can lead to poor generalization and severe catastrophic forgetting when models are adapted to new tasks or domains \citep{ kirkpatrick2017overcoming, li2017learning}. This tension raises a fundamental challenge for continual learning: \emph{how can we obtain the benefits of on-policy learning when only demonstrations are available?}

\begin{figure*}[!t]
    \centering
    \includegraphics[width=\textwidth]{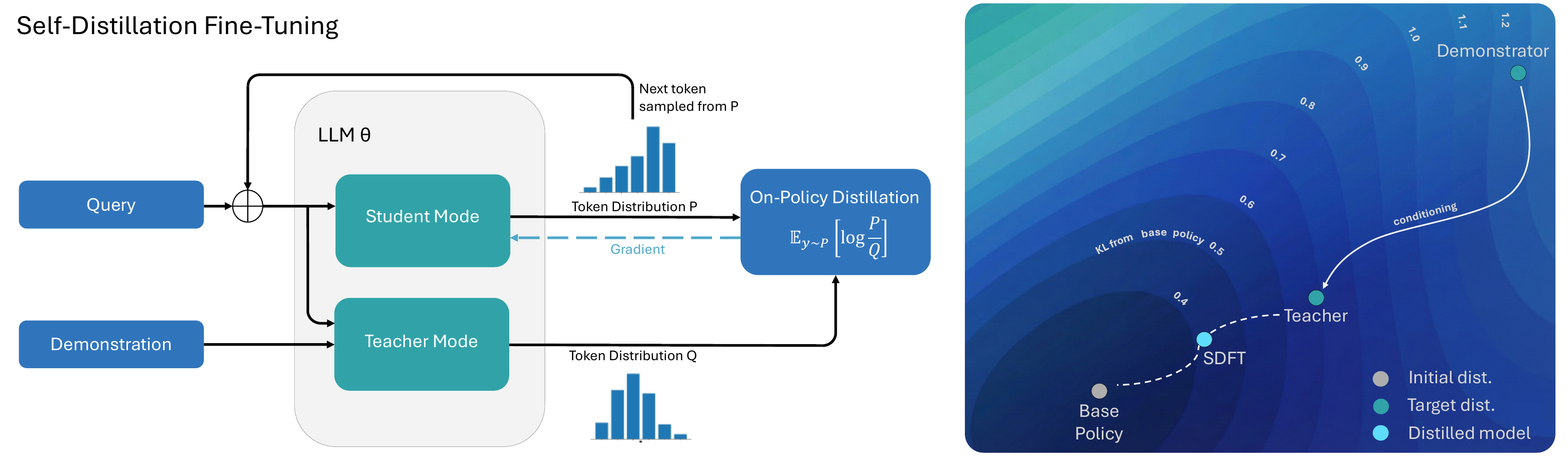}
    \caption{\textbf{(Left)} SDFT leverages a model’s in-context learning ability to generate on-policy training signals. 
    For each query $x$, the model acts in two roles. A student that is conditioned only on the query $P=\pi(\cdot|x)$ and the teacher, which is the same model conditioned on an expert demonstration $c$, producing a demonstration-aware distribution $Q=\pi(\cdot|x,c)$. Training minimizes the reverse KL divergence between the student and teacher, yielding on-policy updates. \textbf{(Right)} Conditioning the model on the expert demonstrations creates a teacher with an output distribution that is substantially closer to the base model, while maintaining the same new-task accuracy.}
    \label{fig:kl_fig}
\end{figure*}

The challenges of off-policy learning can, in principle, be overcome by first learning a reward function from demonstrations (i.e., Inverse Reinforcement Learning or IRL), and then performing on-policy RL \citep{ng2000algorithms, abbeel2004apprenticeship}. 
While IRL is conceptually elegant, effectively recovering rewards typically requires strong priors over the reward structure, which has limited its practical adoption to settings where such assumptions are justified, such as RLHF \citep{ peng2018deepmimic, stiennon2020learning}.

Rather than inferring an explicit reward function, we propose Self-Distillation Fine-Tuning (SDFT), an on-policy distillation \citep{ross2011reduction, agarwal2024policy} framework for learning directly from demonstrations. SDFT relies on the observation that large pretrained models exhibit strong in-context learning—the ability to adapt their behavior when conditioned on examples, without parameter updates \citep{brown2020language}. We exploit this property by using the same model in two roles: a teacher, conditioned on both the task input and an expert demonstration, and a student, conditioned only on the task input. Training distills the teacher’s predictions into the student on trajectories generated by the student itself, yielding on-policy updates that incorporate information from demonstrations without explicit reward inference or offline imitation.

We evaluate SDFT in two continual learning settings: \emph{skill learning}, where demonstrations are used to improve performance on a task, and \emph{knowledge acquisition}, where new information must be incorporated into the model. Across both settings, SDFT provides stable on-policy updates that enable learning while substantially reducing catastrophic forgetting compared to supervised learning. Consistent with prior work on on-policy learning \citep{ross2011reduction, chu2025sft}, SDFT also improves generalization both in-distribution and out-of-distribution, making it beneficial even in settings where retaining prior capabilities is not the primary objective. In a sequential learning experiment involving three distinct skills, SDFT enables a single model to acquire each skill in turn while preserving performance on previously learned skills as well as on unrelated, pre-existing capabilities — demonstrating that continual learning from demonstrations is possible.

\section{Related Work}

\paragraph{Off-policy versus On-policy Learning.} A long line of work highlights the advantages of on-policy learning, i.e., training on trajectories induced by the model itself, over off-policy learning. The seminal result of \citet{ross2011reduction} shows that off-policy imitation learning suffers from compounding errors at inference time, as the learned policy drifts away from the states covered in the demonstrations, errors accumulate rapidly, a failure mode that on-policy algorithms avoid by continually training under their own state distribution. More recent empirical studies reinforce this distinction. Models fine-tuned with on-policy RL have been shown to generalize better beyond the training distribution \citep{agarwal2024policy,han2025general, chu2025sft, li2025unveiling} and transfer more effectively to related tasks \citep{huan2025does} than models trained purely off-policy. In continual learning settings, on-policy updates also reduce catastrophic forgetting when adapting to new tasks \citep{shenfeld2025rl, lai2025reinforcement}. These findings collectively motivate our goal - to enable on-policy learning from demonstrations, thereby retaining the benefits of on-policy RL while avoiding the need for explicit reward engineering.

\paragraph{Inverse Reinforcement Learning.} 
IRL \citep{ng2000algorithms} provides a classical solution to the problem faced in many RL settings: the agent must learn a policy when no explicit reward function is available, only demonstrations. Rather than cloning the expert’s actions, IRL seeks to infer the underlying reward for which those demonstrations would be optimal. This perspective avoids the issues of off-policy imitation learning, since the inferred reward can support on-policy updates \citep{xu2020error}. While this idea has deep theoretical appeal, traditional IRL methods are known not to scale well \citep{lazzati2024does, arora2021survey}.

A common thread across all successful IRL formulations is that they rely on strong structural assumptions to make the reward identifiable. Maximum-entropy IRL assumes that experts follow a soft-optimal Boltzmann policy \citep{ziebart2008maximum, wulfmeier2015maximum}; adversarial IRL methods \citep{ho2016generative} assume that expert and learner trajectories can be distinguished by a classifier; and preference-based IRL methods, such as RLHF \citep{ziegler2019fine, ouyang2022training}, assume access to pairs of positive–negative demonstrations. These priors are essential—without them, IRL is either ill-posed or too expensive to be practical. 
In our approach, rather than imposing an explicit learning reward function, we leverage the model’s in-context learning to extract an on-policy learning signal. 

\paragraph{Context Distillation.} Our method also relates to the growing line of work on context distillation, in which a model conditioned on additional information acts as a teacher for a version of itself without that information \citep{bai2022constitutional, snell2022learning}. Prior approaches typically rely on offline distillation from static contexts, such as few-shot examples or behavioral guidelines, and supervise the student on trajectories drawn from the teacher’s distribution. Our algorithm differs in two important ways. First, the distillation is on-policy: the student is trained under its own induced trajectory distribution, allowing the teacher to correct errors as they arise \citep{ross2011reduction, agarwal2024policy}. Second, the context provided to the teacher is not a fixed prompt prefix but a \textit{specific demonstration chosen for each query}. This dynamic, instance-wise conditioning enables the teacher to express fine-grained task intent, rather than a single global behavioral prior. Together, these differences allow context distillation to function not merely as a form of prompt compression but as an IRL-like mechanism that extracts and transfers the underlying reasoning induced by demonstrations.

\paragraph{Self-Distillation.} Concurrently and independently of our work, two other papers have suggested the Self-Distillation algorithm \cite{zhao2026self, hubotter2026reinforcement}. We see these works as complementary to our own, demonstrating the potential in this algorithm for training LLMs.

\section{Self-Distillation Fine-Tuning}
\label{subsec:algo}

Our approach builds on the framework of student-teacher distillation, where a student model is trained to match the behavior of a teacher model by minimizing the divergence between their output distributions. Traditionally, distillation uses separate models, typically a larger, more capable teacher and a smaller student \citep{hinton2015distilling}. Our key innovation is that we can use the \emph{same} model as both teacher and student by exploiting its in-context learning abilities. Specifically, given a foundation model with policy $\pi$, we construct the teacher by conditioning it on expert demonstrations: $\pi(\cdot|x, c)$, where $x$ is the task prompt and $c$ is a demonstration. The student is simply the base model without this conditioning $\pi_\theta(\cdot|x)$. 

To construct the teacher for a given prompt $x$, we condition the model on both the prompt and a demonstration using the following simple prompt:

\begin{verbatim}
<Question>
This is an example for a response to the question:
<Demonstration>
Now answer with a response of your own, including the thinking process: 
\end{verbatim}

We find that this prompt is sufficient to prevent the policy from outputting $c$ verbatim and instead elicits a response that reflects the model's understanding of the intent behind the demonstration, leveraging its in-context learning capabilities. See subsection \ref{subsec:assumption} for further analysis of the conditioned policy's outputs. 

As mentioned before, we hypothesize that on-policy learning is necessary for continual learning; therefore, we train the student using on-policy distillation from the teacher\footnote{Even without the continual learning constraints, learning from a teacher with privileged information requires on-policy learning \citep{swamy2022sequence}.}. For every prompt $x$, our algorithm, SDFT, samples responses from the student policy $y\sim\pi_\theta(\cdot|x)$ and minimizes the reverse Kullback-Leibler (KL) divergence between the student and the teacher distributions:
\begin{equation}
    \mathcal{L}(\theta) = D_{KL}\left(\pi_\theta(\cdot|x) \parallel \pi(\cdot|x, c)\right) = \mathbb E_{y\sim\pi_\theta(y|x)}\left[\log \frac{\pi_\theta(y|x)}{\pi(y|x, c)}\right]
\end{equation}
Leveraging the autoregressive nature of the model, we decompose this objective into a token-level loss (see \citet{tang2025few} for a derivation)
and take the gradient with respect to the student parameters $\theta$ while treating the teacher distribution as fixed. This results in the following gradient estimator:
\begin{equation}
\label{eq:gradient}
    \nabla_\theta \mathcal{L} (\theta) =
    \mathbb{E}_{y \sim \pi_\theta} \left[ \sum_t \sum_{y_t\in \mathcal{V}} \pi_\theta(y_t|y_{<t}, x)\log \frac{\pi_\theta(y_t|y_{<t}, x)}{\pi(y_t|y_{<t}, x, c)} \nabla_\theta \log \pi_\theta(y_t|y_{<t}, x) \right]
\end{equation} 
where $\mathcal{V}$ is the token vocabulary.

\textbf{Practical Implementation. }There are a few critical components to the practical implementation of SDFT. The first is the parameterization of the teacher model used to compute the likelihood ratios. 
Appendix \ref{app:ema_ablation} includes an ablation regarding this design choice. Unless mentioned otherwise, we use an exponential moving average (EMA) of the student parameters for the teacher. Another one is the logit-level loss taken between the teacher and the students \cite{agarwal2024policy}. Although the theory points to Reverse KL as a suitable loss, we found in practice that Forward KL yields the best performance. A full detailed description of our algorithm can be found in Algorithm \ref{alg:sdft} in the appendix. 

\subsection{Self-Distillation as Inverse RL}

Although we present our algorithm from a student-teacher distillation perspective, it can also be interpreted in the IRL framework, where it maximizes an implicit reward function. In the following section, we formally show that our self-distillation objective is mathematically equivalent to maximizing an implicit reward function defined by the expert demonstrations and the model's in-context learning capabilities. 

We begin with the standard formulation of trust-region-regularized reinforcement learning \cite{schulman2015trust}, where the policy update in step $k+1$ is constrained to stay close to the current policy $\pi_k$:
\begin{equation}
\label{eq:constraint}
\begin{split}
\pi_{k+1} = \max_{\pi} \mathbb{E}_{y \sim \pi} [r(y,x)]-\beta D_{\mathrm{KL}}(\pi(\cdot|x)\|\pi_k(\cdot|x))
\end{split}
\end{equation}

For this objective, the optimal policy $\pi_{k+1}^*$ takes the known closed-form expression of a tilted distribution \citep{korbak2022rl, rafailov2023direct}:
\begin{equation*}
\pi_{k+1}^*(y|x) \propto \pi_k(y|x)\,\exp(\frac{1}{\beta}\,r(y,x))
\end{equation*}
Rearranging this equation allows us to express the underlying reward as a function of the divergence between the optimal and previous policies:
\begin{equation*}
r(y,x) = \beta\left[\log \pi_{k+1}^*(y|x) - \log \pi_k(y|x)\right] + C
\end{equation*}

In a standard IRL setting, $\pi_{k+1}^*$ is unknown. However, our key idea is that the model's own in-context learning capabilities provide a robust approximation of this optimal policy. We introduce our \textit{In-Context Assumption} - given a demonstration $c$, the model conditioned on $c$ approximates the optimal next policy.
\begin{equation}
\label{eq:assumption}
\pi_{k+1}^*(y|x) \approx \pi(y|x, c)
\end{equation}
This substitution posits that the behavioral shift induced by observing a demonstration reflects the expert's true intent. Substituting this into Eq. (6), we derive an intrinsic reward function:
\begin{equation}
\label{eq:reward}
r(y,x,c) = \log \pi(y|x,c) - \log \pi_k(y|x)
\end{equation}
We drop $\beta$ and $C$ since linear transformations of reward do not affect the optimal policy \citep{sutton1998reinforcement}. 
While this defines a trajectory-level reward, our model has an autoregressive structure. Therefore, we decompose the reward into token-level rewards $r_t$ via token-level probabilities. We define the instantaneous reward as the immediate log-probability change:
\begin{equation*}
r_t(y_t \,|\, y_{<t}, x,c) = \log \frac{\pi(y_t \,|\, y_{<t}, x, c)}{\pi_k(y_t \,|\, y_{<t}, x)},
\end{equation*} and indeed for all $y$, we have $\sum_t r_t(y_t \,|\, y_{<t}, x,c)=r(y,x,c)$.
Finally, we demonstrate that optimizing the policy with respect to this reward is equivalent to the reverse-KL distillation used in our method. The policy gradient under the current policy $\pi_k$ is:
\begin{equation*}
\nabla_\theta J(\pi_k)
    = \mathbb{E}_{y \sim \pi_k}
    \left[ r(y,x,c) \nabla_\theta\log \pi_k(y|x) \right]
\end{equation*}
Substituting our derived reward from Equation~\ref{eq:reward}:
\begin{equation}
\nabla_\theta J(\pi_k)
    = \mathbb{E}_{y \sim \pi_k}
    \left[\log \frac{\pi(y|x,c)}{\pi_k(y|x)}\nabla_\theta \log \pi_k(y|x)\right]
\end{equation}

We observe that this is equivalent in expectation to the gradient of the reverse KL divergence $D_{\mathrm{KL}}(\pi_k(\cdot|x) \| \pi(\cdot|x,c))$ in Equation~\ref{eq:gradient}.
Thus, our method can be viewed as an on-policy RL algorithm that maximizes rewards inferred by comparing the student's current behavior to its own ``wiser,'' demonstration-aware counterpart.

\begin{figure*}[t]
    \centering
    \begin{subfigure}[t]{0.5\textwidth}
        \centering
        \includegraphics[width=\textwidth]{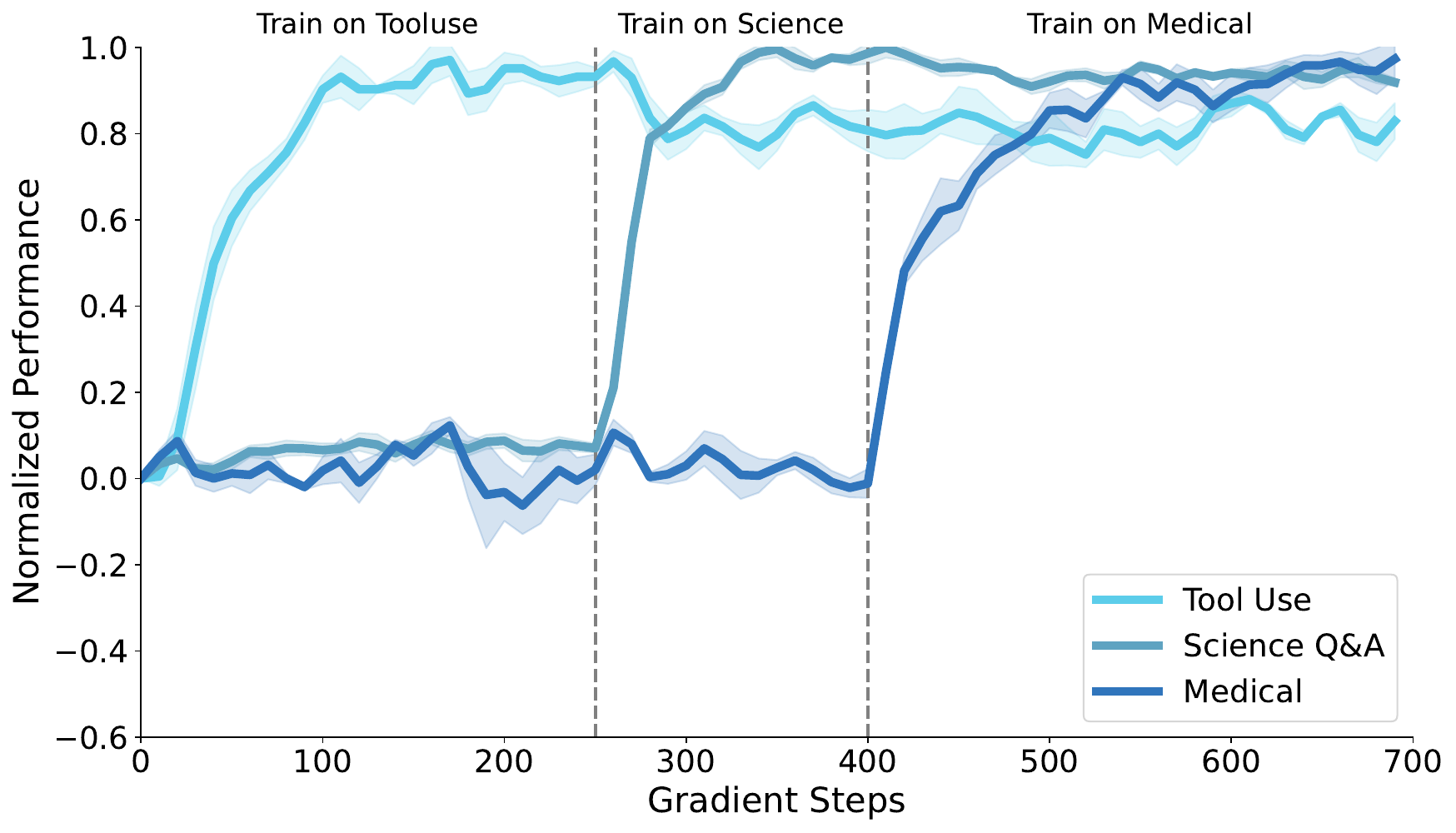}
        \caption{SDFT}
    \end{subfigure}\hfill
    \begin{subfigure}[t]{0.5\textwidth}
        \centering
        \includegraphics[width=\textwidth]{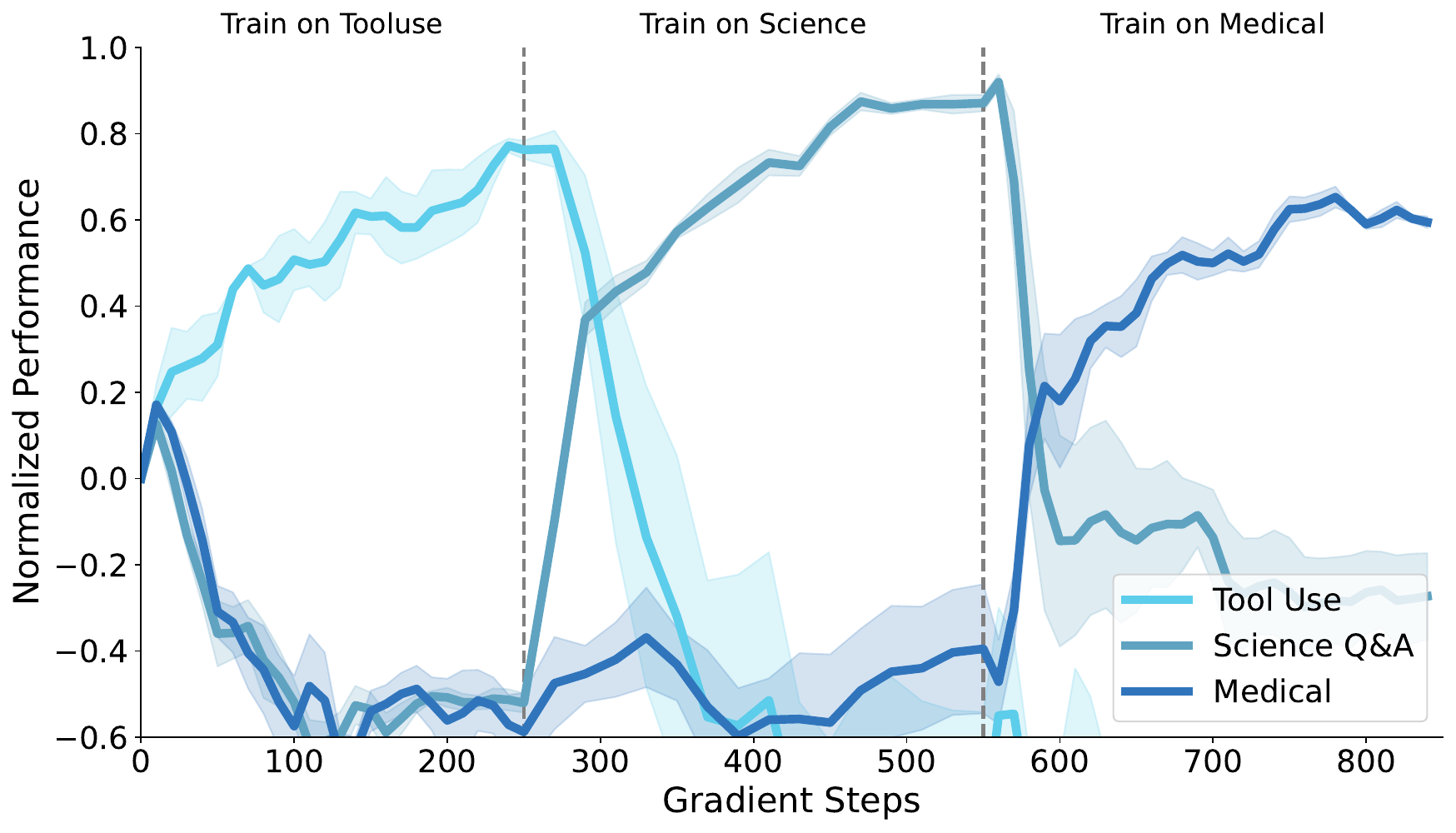}
        \caption{SFT}
    \end{subfigure}
    \caption{In a challenging continual learning experiment, where one model is trained sequentially on three different tasks, SDFT is able to learn each one while retaining performance on the others. In contrast, SFT performance on each task drops once it starts learning the next one. Performance is linearly normalized such that 0 corresponds to the base model accuracy on each one of the tasks, and 1 to the maximum accuracy obtained across both algorithms.}
    \label{fig:seq_learning}
\end{figure*}

\subsection{Validating the ICL Assumption}
\label{subsec:assumption}
The core hypothesis of SDFT can be seen as the assumption in Equation~\ref{eq:assumption}, which states that a model conditioned on an expert demonstration behaves like the (unknown) optimal policy for that task $\pi_{k+1}^*(y|x) \approx \pi(y|x,c)$ and therefore it can be a good teacher. The quality of this approximation depends on 2 conditions:
\begin{enumerate}[leftmargin=*]
    \item \emph{Optimality}: The teacher’s expected reward must match that of the unknown optimal policy:
    $$\mathbb{E}_{y \sim \pi(y|x,c)}[r(y,x)] \approx\mathbb{E}_{y \sim \pi_{k+1}^*}[r(y,x)]$$
    In other words, samples drawn from the demonstration-conditioned policy should achieve near-maximal reward on the task.
    \item \emph{Minimal Deviation}: Due to the trust-region regularization in Equation~\ref{eq:constraint}, the optimal policy $\pi_{k+1}^*(y|x)$ is the one closest to the current model among all the ones that maximize reward. Thus, we require:
    $$D_{\mathrm{KL}}\left(\pi(\cdot|x,c)||\pi_k(\cdot|x)\right)\approx D_{\mathrm{KL}}\left(\pi_{k+1}^*(\cdot|x)||\pi_k(\cdot|x)\right)$$
    That is, among all policies that achieve optimal reward, the teacher should be close, in the KL sense, to the $\pi_k$.
\end{enumerate}

The second requirement, remaining close to the current policy, is crucial for practical viability. If the demonstration-conditioned teacher simply mimicked the example verbatim, it would deviate substantially from the base model, losing the benefits of on-policy learning. What makes the teacher valuable is that it produces new, task-appropriate behavior while remaining anchored to the base model. Moreover, prior work shows that distributions close to the pretrained distribution suffer significantly less catastrophic forgetting and better preserve general capabilities \citep{shenfeld2025rl, chen2025retaining}.

\textbf{Empirical Validation.} While we cannot verify these conditions theoretically, we evaluate each empirically. We use the Qwen-2.5-7B-Instruct model \citep{hui2024qwen2} as the base policy and the ToolAlpaca dataset \citep{tang2023toolalpaca}. In this benchmark, the model receives a tool-API specification and a user request, and must identify the correct tool call. Without demonstrations, the base model solves only 42\% of examples. When provided with the appropriate demonstration $c$ for each prompt $x$, the teacher achieves a 100\% success rate.  To further test reward proximity, we manually inspected 50 teacher reasoning traces. In all cases, not only were the final tool calls correct, but the intermediate chain-of-thought was valid and semantically grounded. This suggests that the teacher is reconstructing a correct reasoning process rather than merely copying the expert output. These observations provide evidence for the first requirement, that the demonstration-conditioned model behaves as an optimal policy.

To verify the second requirement, we measure the KL divergence to the base policy  $D_{\mathrm{KL}}(\pi\|\pi_0)$ as a proxy for the distance to the policy during training  $\pi_k$. We compare this divergence for both the SFT model trained on demonstrations and the demonstration-conditioned teacher. As shown in Figure~\ref{fig:kl_fig} (right panel), the SFT model deviates substantially from the base model (1.26 nats), whereas the teacher remains significantly closer (0.68 nats)—nearly half the divergence. This validates that the teacher produces high-quality outputs while maintaining proximity to the base policy, precisely the balance required by the trust-region formulation.

\section{Experiments}

\subsection{Experimental Setting}
\label{subsec:exp_setting}
We evaluate our method in two settings that reflect common forms of post-training adaptation: \emph{Skill Learning} and \emph{Knowledge Acquisition}. These correspond to improving performance on a new task, and integrating novel factual information into a pretrained model.

In \emph{Skill Learning}, we study whether a pretrained LLM with broad capabilities can acquire a new, narrowly defined skill without degrading its existing abilities. We choose to experiment with tasks the models had not been explicitly fine-tuned on (unlike Math or Coding) to show the benefits of continual learning. Therefore, we test our method on three domains:

\begin{itemize}[leftmargin=*]
    \item \textit{Science Q\&A}: Undergraduate-level scientific reasoning, using the Chemistry L-3 subset of SciKnowEval \citep{feng2024sciknoweval}.
    \item \textit{Tool Use}: Mapping a tool-API specification and user request to the correct tool call, using ToolAlpaca \citep{tang2023toolalpaca}.
    \item \textit{Medical}: Clinical reasoning questions, with training data from stage 1 of the HuatuoGPT-o1 pipeline and evaluation from stage 2 \citep{chen2024huatuogpt}.
\end{itemize}

In \emph{Knowledge Acquisition}, the objective is different: the model must integrate genuinely new factual content not present in its pretraining data. We construct a corpus of Wikipedia articles describing natural disasters that occurred in 2025 (after the training knowledge cutoff), totaling approximately 200K tokens. Following \citet{mecklenburg2024injecting}, we generate question–answer pairs about these articles, yielding an SFT dataset roughly 5× larger than the source corpus. These questions probe factual content such as “which regions were affected by the 2025 Myanmar earthquake?”. 

This setting tests whether the model can absorb newly injected knowledge rather than merely improving skills it already has.

\paragraph{Evaluation.}

For each task, we evaluate along two primary axes:\looseness=-1
\begin{itemize}[leftmargin=*]
\item \emph{In-Distribution Accuracy:} Accuracy on held-out test data for the newly introduced task. For Knowledge Acquisition, we use two variants: (1) All details correct (Strict Accuracy). (2) The answer contains correct information and no incorrect statements (Lenient Accuracy).
\item \emph{Previous Capabilities:} Performance on a suite of established benchmarks that probe general reasoning and world knowledge: HellaSwag \citep{zellers2019hellaswag}, TruthfulQA \citep{lin2021truthfulqa}, MMLU \citep{hendrycks2020measuring}, IFEval \citep{zhou2023instruction}, Winogrande \citep{sakaguchi2021winogrande}, and HumanEval \citep{chen2021evaluating}. We report the average performance across these datasets as a measure of catastrophic forgetting. 
\end{itemize}

For the \emph{Knowledge Acquisition} setting, we include a third metric:
\begin{itemize}[leftmargin=*]
    \item \emph{Out-of-Distribution Accuracy:} “Indirect” questions whose answers depend on the injected knowledge but do not directly reference it (e.g., “Which countries required international humanitarian aid in 2025?”). This measures whether the new information has been properly integrated into the model’s internal memory rather than memorized in a narrow form.
\end{itemize}

\begin{figure*}[t]
    \centering

    \includegraphics[width=\textwidth]{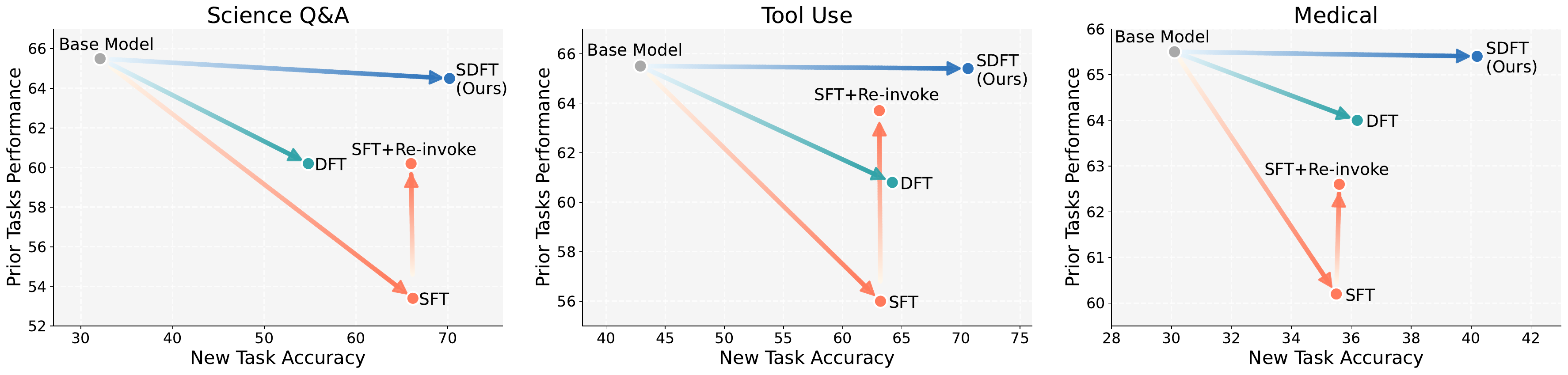}
    \caption{Performance trade-offs between new task accuracy and retention of prior capabilities. Each point represents a trained model, with the top-right indicating ideal performance (high accuracy on both new and previous tasks). SDFT consistently achieves superior Pareto efficiency compared to baselines across all three skill learning tasks.}
    
    \label{fig:skills}
\end{figure*}

\paragraph{Baselines.} In the \textit{Skill Learning} setting, we compare our method to standard \textbf{SFT} and to \textbf{DFT} \citep{wu2025generalization}, which uses importance sampling to treat the offline dataset as on-policy samples. We also include the recently proposed "\textbf{Re-invocation}" method \citep{lu2025onpolicydistillation}, which performs additional on-policy distillation from the base policy on general-purpose prompts after SFT to restore prior capabilities. 

In the \textit{Knowledge Acquisition} setting, we compare our method to \textbf{CPT} (Continual Pre-Training), which trains directly on the text corpus using next token prediction loss and \textbf{SFT}, which trains on the question-answer pairs. In addition, we also compare with pure ICL methods. Because the full corpus exceeds the model’s context window, we evaluate \textbf{RAG} with an oracle retriever that always provides the correct article for each question.

Unless otherwise noted, all experiments were performed on the Qwen2.5-7B-Instruct model. For each baseline, we perform a hyperparameter sweep and report results for the model achieving the highest validation performance on the target task. Full datasets, hyperparameters, and training protocols are provided in Appendix~\ref{app:hps}.

\subsection{On-policy learning leads to better generalization}

\begin{wraptable}{r}{0.45\textwidth}
    \vspace{-0.5em}
    \centering
    \begin{tabular}{@{\extracolsep{\fill}}lccc@{}}
      \toprule
       & Accuracy  & Accuracy  & OOD  \\
       & (strict) & (lenient) & Accuracy \\
      \midrule
      Base       & 0   & 0   & 0   \\
      Oracle RAG & 91   & 100 & 100 \\
      \midrule
      CPT        & 9   & 37  & 7   \\
      SFT        & 80  & 95  & 80  \\
      SDFT (Ours)& \textbf{89} & \textbf{100} & \textbf{98} \\
      \bottomrule
    \end{tabular}
    \caption{SDFT effectively integrates new factual knowledge, thus achieving better accuracy both in- and out-of-distribution.}
    \label{tab:KI}
    \end{wraptable}

Prior work has shown that on policy learning achieves better in-distribution performance than SFT \citep{ross2011reduction}, as well as superior out-of-distribution generalization \citep{chu2025sft}. We investigate whether these advantages also arise in our on-policy distillation framework. For that, we measure performance on test set on all our training tasks, as well as OOD generalization in the Knowledge Acquisition setting.

\paragraph{Results.} Results for \emph{Skill Learning}, as shown in Figure \ref{fig:skills}, indicate that our method achieves higher new-task accuracy than SFT, which represents better in-distribution generalization. We attribute these gains to the fact that off-policy learning trains only on expert-induced trajectories; errors at test-time can push the policy into unseen states, causing compounding errors. On-policy imitation learning avoids this mismatch by training on the state distribution induced by the learned policy itself \citep{ross2011reduction}.

The results for \textit{Knowledge Acquisition} appear in Table~\ref{tab:KI}. Since the new knowledge was not included in the base model's training, it cannot answer any of the questions correctly. Consistent with earlier observations \citep{mecklenburg2024injecting}, continual pretraining performs poorly. SFT on questions improves performance substantially but still lags behind our SDFT. On strict accuracy, it reaches 80\% while our on-policy method achieves 89\% and nearly closes the gap to the oracle RAG model. The advantage becomes even clearer on out-of-distribution questions, where our method achieves close to perfect accuracy, while SFT’s performance remains low. This disparity underscores a key limitation of SFT: it teaches the model to reproduce specific answers but does not reliably incorporate the underlying facts into the model’s broader knowledge base.

Finally, with on-policy RL there is a concern for superficial improvements through entropy reduction rather than acquisition of new behaviors \citep{yue2025does, wu2025invisible}. To ensure our gains are not merely due to distributional sharpening, we evaluate pass@$k$ for $k$ up to 128 in the Skill Learning Setting. As shown in Figure~\ref{fig:pass_at_k} (right), the performance gains over both the base model and SFT persist uniformly across all $k$. This indicates that the improvements reflect genuine skill acquisition rather than entropy collapse.


\subsection{Learning without forgetting}

A central claim of SDFT is that, due to its on-policy nature, it can acquire new skills while mitigating catastrophic forgetting. To test this, we perform the following experiments:
\begin{enumerate}[leftmargin=*]
    \item \textbf{Single Task Learning.} A convenient case study for continual learning is fine-tuning a model on a single task. Using the \emph{Skill Learning} setting, we compare the broad capabilities of our models before and after training on each task.
    \item \textbf{Multi-Task Continual Learning.} We investigate a more complex continual learning experiment in which a single model is trained sequentially on each task. The goal here is to measure catastrophic forgetting over longer training and to see whether the model retains the capabilities it learned at each stage of training.
\end{enumerate}. 
\vspace{-20pt}
\paragraph{Results.}
The results for single-task training, presented in Figure \ref{fig:skills}, show that our method is the only approach to improve performance on the new task without significant degradation in prior capabilities. In contrast, standard SFT produces substantial catastrophic forgetting across all evaluated benchmarks. Augmenting SFT with the “re-invoke’’ procedure partially restores lost abilities but does not recover the base model’s full capabilities. DFT, which performs approximate on-policy updates, exhibits reduced forgetting relative to SFT but still results in noticeable degradation. For the breakdown of the score over prior tasks, see Table \ref{tab:skills}.

We now turn to the more challenging setting of long-horizon continual learning, where a single model is trained sequentially on all three skills. Figure \ref{fig:seq_learning} shows that SDFT enables stable accumulation of skills over time. As training progresses, the model improves on each newly introduced task while maintaining performance on previously learned ones. In contrast, SFT exhibits severe interference—performance on earlier skills rapidly degrades once training shifts to a new task, resulting in oscillatory behavior rather than cumulative learning. 
These results demonstrate that SDFT supports true continual learning, allowing a single model to incrementally acquire multiple skills without catastrophic forgetting.

\subsection{Effect of model size}

\begin{figure*}[t]
    \centering
    \begin{subfigure}[t]{0.48\textwidth}
        \centering
        \includegraphics[width=\textwidth]{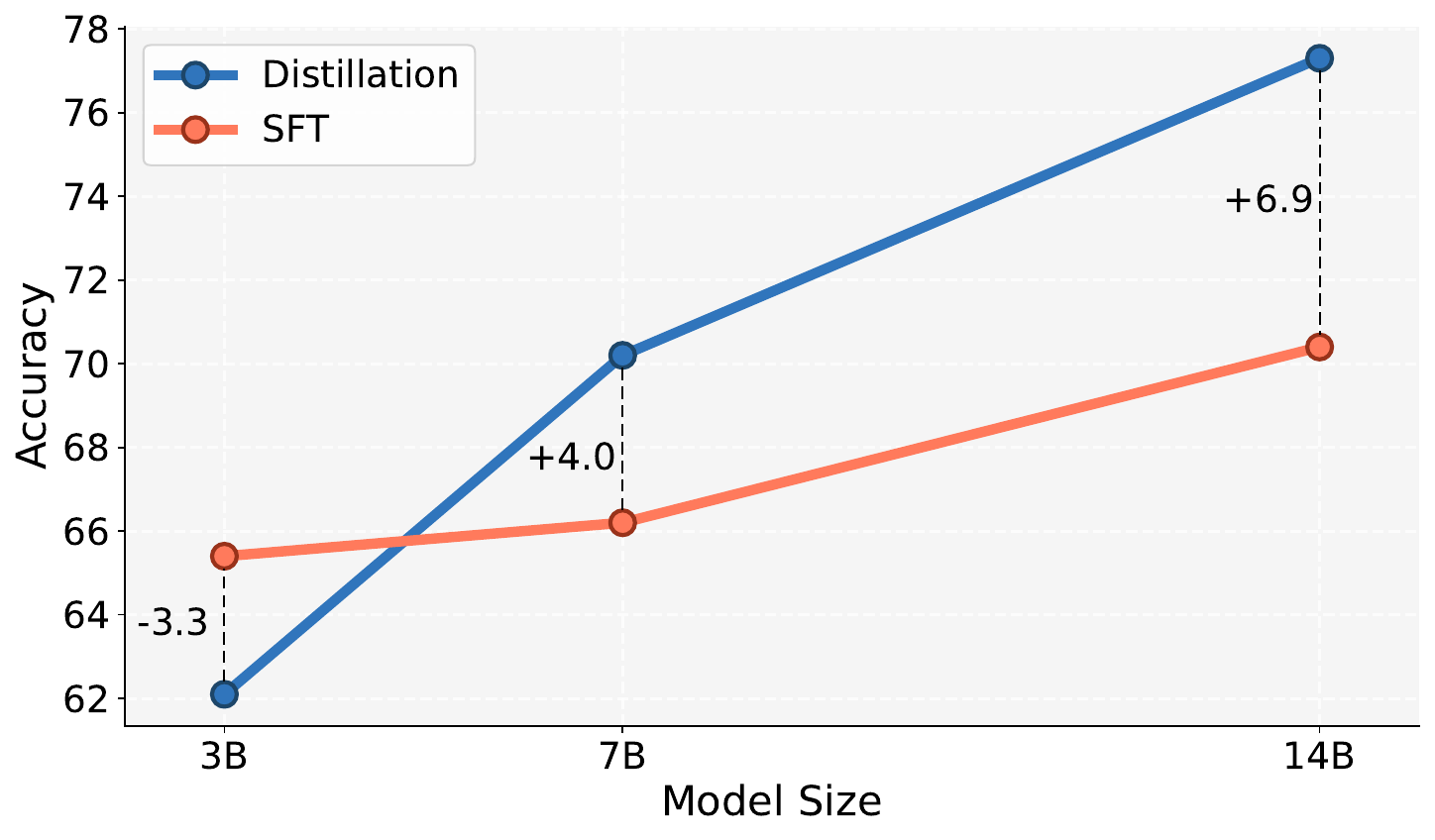}
    \end{subfigure}\hfill
    \begin{subfigure}[t]{0.48\textwidth}
        \centering
        \includegraphics[width=\textwidth]{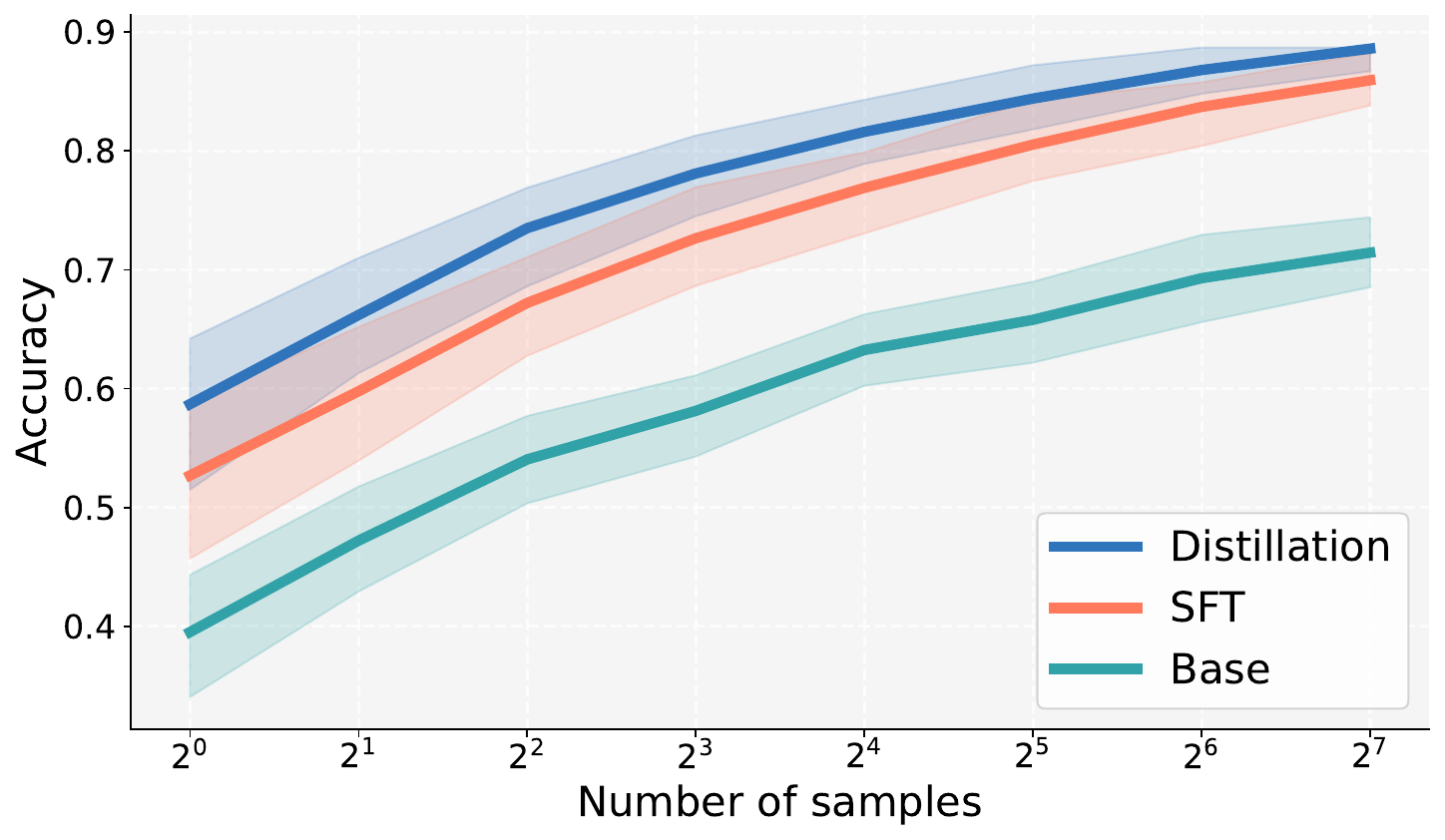}
    \end{subfigure}
    \caption{\textbf{(Left)} SDFT benefits from model scale. Performance gap between SDFT and SFT on the Science Q\&A task increases with model size, as larger models have stronger in-context learning capabilities. \textbf{(Right)} SDFT improves pass@k across various k, indicating genuine skill acquisition rather than entropy collapse.}
    \label{fig:pass_at_k}
\end{figure*}

Our method relies fundamentally on the model’s in-context learning ability. The teacher signal comes from the model conditioned on demonstrations, and the quality of that signal depends on how well the model can interpret and extrapolate from those examples. This suggests that larger models, whose in-context learning abilities are known to improve with scale \citep{brown2020language}, should yield stronger teacher policies and therefore better SDFT updates. To test this hypothesis, we conduct a scaling experiment using several sizes from the Qwen 2.5 family \citep{hui2024qwen2}, evaluating each on the Science Q\&A task.

Figure~\ref{fig:pass_at_k} (left) illustrates a clear trend. At small scales, such as the 3B variant, the model’s in-context learning is too weak to provide meaningful teacher guidance, and performance lags behind standard SFT. However, as we increase the size, the gains from our method grow consistently. The 7B model achieves a four-point improvement over SFT, and the 14B model widens the margin to seven points. This monotonic improvement suggests that the effectiveness of our algorithm is tightly coupled to the model’s ability to perform in-context reasoning, and, as larger models continue to exhibit stronger ICL behavior, our approach is likely to become even more advantageous at larger scales.

\subsection{Training reasoning models without reasoning data}

A major practical challenge in post-training reasoning models is the construction of high-quality supervision. Supervised fine-tuning for reasoning models typically requires access to intermediate reasoning traces, which are expensive to collect from human annotators and often unavailable from closed-source models that expose only final answers. As a result, many real-world datasets provide only the final answer, without the full chain of thought.

Naively applying SFT on such data can be harmful for reasoning-capable models. Because SFT directly matches the target responses, when demonstrations contain only short answers or abbreviated reasoning, this suppresses the model’s existing long chain-of-thought behavior. For example, consider a model that reliably produces long chains of thought, but is post-trained using demonstrations that provide only concise solutions. Direct SFT penalizes long reasoning traces in favor of short outputs that better match the supervision, causing a collapse in reasoning depth. We hypothesize that our on-policy self-distillation framework avoids this failure mode. Because the student policy is trained to match the outputs of a demonstration-conditioned teacher derived from the same model, the supervision preserves the model’s internal reasoning style even when the external data contains only final answers. In effect, the teacher induces a reasoning-consistent target distribution, allowing the model to adapt without collapsing its reasoning behavior.

\begin{wraptable}{r}{0.5\textwidth}
\vspace{-0.5em}
\centering
\begin{tabular}{lcc}
\toprule
                & \multicolumn{1}{l}{Accuracy} & \multicolumn{1}{l}{Avg. \# of tokens} \\ \midrule
Olmo-3-7B-Think & 31.2   & 4612  \\
+ SFT           & 23.5   & 3273   \\
+ SDFT (Ours)  & 43.7   & 4180  \\ 
\bottomrule
\end{tabular}
\caption{Training reasoning models with answer-only supervision. SFT degrades both task performance and general reasoning behavior (indicated by shortened responses). SDFT avoids this collapse by learning from a demonstration-conditioned teacher rather directly from the demonstrations.}
\label{tab:thinking}
\vspace{-2em}
\end{wraptable}

We evaluate this hypothesis using Olmo-3-7B-Think \citep{olmo2025olmo} and fine-tune it on the medical task described earlier. Importantly, this dataset contains no explicit chain-of-thought annotations. We compare standard SFT against our method, measuring both task accuracy and the average number of generated tokens, which serves as a proxy for retained reasoning depth.

\paragraph{Results.}
Table~\ref{tab:thinking} presents the results. Standard SFT substantially degrades performance, reducing accuracy from 31.2\% to 23.5\% and sharply shortening responses, indicating a collapse in reasoning behavior. In contrast, our method significantly improves accuracy, reaching 43.7\%. These results demonstrate that our approach enables effective task adaptation for reasoning models even in the absence of explicit reasoning data.

\subsection{What drives the improvement in performance?}
\label{subsec:ablation}

Our method combines two ingredients: a demonstration-conditioned teacher policy and an on-policy distillation objective. In Subsection~\ref{subsec:assumption}, we validated that the conditioned model constitutes a high-quality teacher—producing correct outputs while remaining close to the base policy. A natural question then arises: \emph{if such a teacher already exists, is on-policy learning necessary, or would standard distillation suffice?}

To isolate the source of the performance gains, we compare our full algorithm against two alternative ways of leveraging the same teacher: (1) SFT from the teacher, where the student is trained offline to imitate samples generated by the teacher \citep{yang2024self, cao2025infiniteicl}. (2) Offline distillation from the teacher, where the student minimizes a KL loss on a fixed dataset of teacher-generated outputs \citep{padmanabhan2023propagating, mitra2025semantic}. We use the Tool Use task for the comparison. 

\begin{wrapfigure}{r}{0.5\textwidth}
    \centering
    \includegraphics[width=0.5\textwidth]{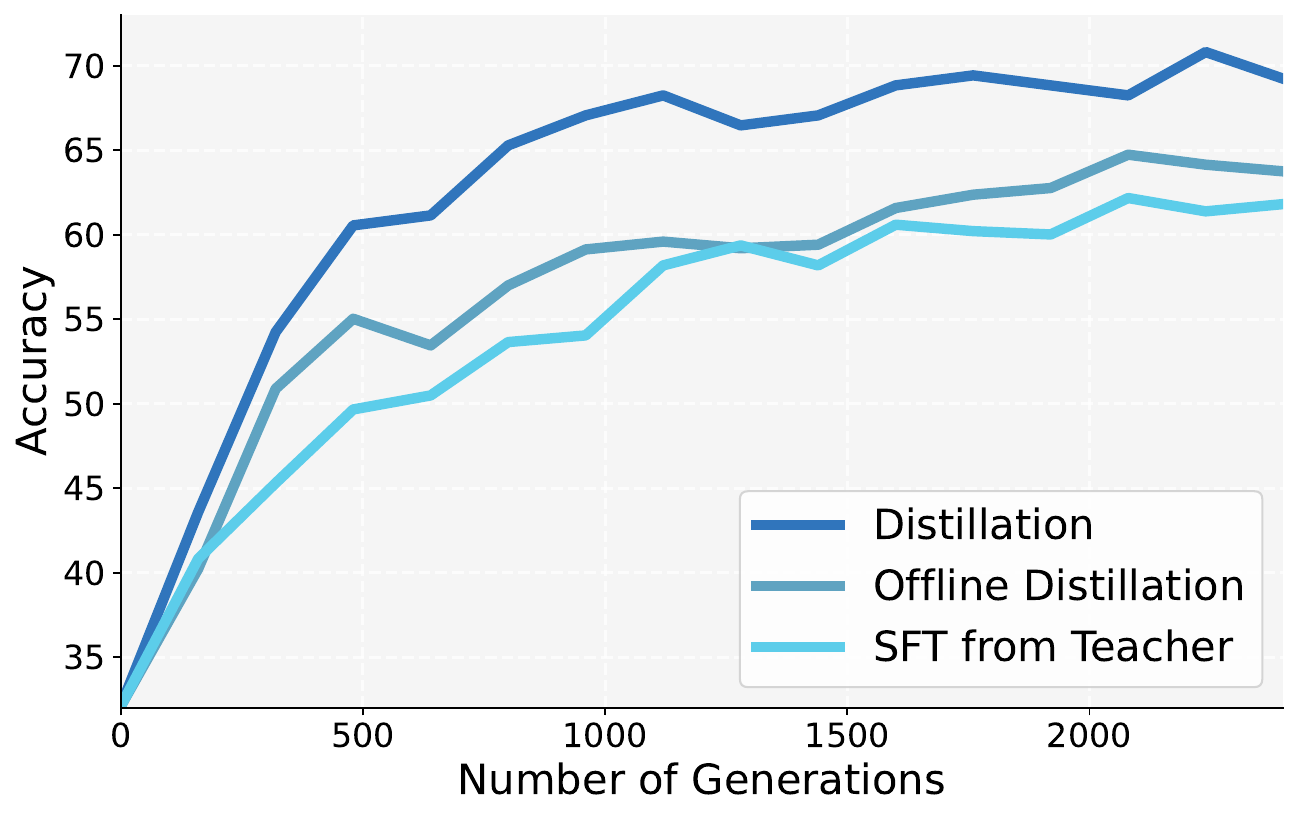}
    \vspace{-3ex}
    \caption{On-policy learning is essential for performance gains. While offline distillation from the improves over standard SFT, it consistently underperforms on-policy SDFT, demonstrating that the benefits cannot be attributed solely to teacher quality.}
    \vspace{-10ex}
    \label{fig:ablation}
\end{wrapfigure}

\paragraph{Results.}
As shown in Figure~\ref{fig:ablation}, neither form of offline distillation matches the performance of our on-policy approach. While distillation from the teacher improves over standard SFT, it consistently underperforms our method. This gap indicates that the benefits of SDFT cannot be attributed solely to the quality of the teacher and further highlights the importance of on-policy learning.

\section{Discussion and Limitations}
\paragraph{Relationship to on-policy RL.}
SDFT is not an alternative to on-policy RL, but addresses a complementary learning regime. The two methods operate under different supervision assumptions: SDFT applies when learning from expert demonstrations without access to an explicit reward function, whereas on-policy RL assumes a reward signal and optimizes expected return through exploration. 

Importantly, the two approaches can be naturally combined. As shown in Figure~\ref{fig:pass_at_k}, SDFT consistently improves pass@$k$ across a wide range of $k$, indicating that it increases the diversity and quality of high-probability generations. This suggests that SDFT can serve as an effective initialization for subsequent RL fine-tuning, providing a stronger starting policy.

If one nevertheless compares the training dynamics, a practical advantage of SDFT is its efficiency. Our method requires only a single on-policy generation per prompt, whereas many on-policy RL methods, such as GRPO, rely on group-based sampling to estimate relative advantages, substantially increasing generation cost. Moreover, our supervision is provided at the token level (or even the logit level), providing denser credit assignment than the trajectory-level advantage used in GRPO.

\paragraph{Computational Costs.} A practical consideration when adopting SDFT is that, unlike standard SFT, it requires generating on-policy rollouts during training. In our experiments, this results in approximately 2.5× the computational cost in FLOPs and roughly 4× the wall-clock training time compared to SFT. However, this additional cost must be viewed in context: many existing continual learning approaches, such as Re-invoke, require sequential training stages—first performing SFT, then conducting additional on-policy training to restore degraded capabilities. When accounting for this multi-stage process, SDFT can actually reduce total training time while simultaneously achieving better performance on the new tasks. 

\paragraph{Learned Artifacts.} A subtle failure mode of our approach is that the student can inherit spurious linguistic patterns from the teacher. Because the teacher is conditioned on demonstrations or text passages, it may produce responses prefaced with phrases like "Based on the text..." or "Following the example..." The student, although receiving no such context, sometimes nevertheless reproduces these markers, having learned them as part of the teacher's output distribution. Empirically, we find that masking the loss over the first few tokens during training effectively suppresses these artifacts without harming downstream accuracy. While this workaround is effective in practice, it is fundamentally a heuristic fix. A more principled solution remains an open problem.

\paragraph{Requirements for Model Capability.} The effectiveness of SDFT depends critically on the base model's in-context learning capabilities. As demonstrated in our scaling experiments (Figure 4a), smaller models with weak ICL abilities fail to provide meaningful teacher signals. Additionally, the fundamental design of our algorithm imposes constraints on the types of adaptations it can support. SDFT excels at acquiring new skills or knowledge while preserving existing capabilities, but struggles when the desired behavior requires a fundamental shift in the model's generation patterns. For instance, we found that transforming a non-reasoning model into one that produces explicit chain-of-thought traces proved difficult. Future work should explore modified objectives or prompting techniques to enable more aggressive behavioral changes when necessary.

\paragraph{Future Work.} Several promising directions remain for extending SDFT. First, our method can be naturally combined with on-policy RL, either sequentially by using SDFT as an initialization before reward-based fine-tuning, or simultaneously by blending demonstration-based and reward-based learning signals. Second, although on-policy learning substantially reduces catastrophic forgetting compared to off-policy methods, some degradation of prior capabilities remains. Developing complementary techniques that further minimize forgetting represents an important avenue for future research. Finally, while we focus on expert demonstrations, extending SDFT to learn from non-expert or noisy demonstrations, or from unstructured data such as user conversations, would broaden its applicability to real-world continual learning settings where high-quality supervision is scarce.

\section*{Author Contributions}
\textbf{Idan Shenfeld} Lead the project and contributed to all aspects of experiments and writing.

\textbf{Mehul Damani} supported this project, made significant
contributions to the design of experiments in discussions and gave valuable advice on presentation and writing. 

\textbf{Jonas Hübotter} supported this project, made significant contributions to the core algorithmic design, and experimental setting. Gave valuable advice on writing.

\textbf{Pulkit Agrawal} co-developed the project direction, advised IS, and played a significant role in paper writing.

\section*{Acknowledgment}
We want to express our gratitude to Nitish Dashora, Jyo Pari, Isha Puri, Zhang-Wei Hong, and members of the Improbable AI lab for the helpful discussion on the paper. We also thank Nofit Segal, Dan Haramati, Yael Vinker, and Assaf Ben-Kish for their support. We are grateful to MIT Supercloud
and the Lincoln Laboratory Supercomputing Center for providing HPC resources. The research was supported in part by Hyundai Motor Company, Google, Amazon and MIT-IBM Watson AI Lab. The research was sponsored
by the Army Research Office and was accomplished under
Grant Number W911NF-21-1-0328 and an ARO MURI under Grant Number W911NF-23-1-0277. The research was also sponsored by the Office of Naval Research and was accomplished under Grant Number N00014-22-1-2740. Research was also sponsored by the Department of the Air Force Artificial Intelligence Accelerator and was accomplished under Cooperative Agreement Number FA8750-19-2-1000. The views and conclusions contained in this document are those of the authors and should not be interpreted as representing the official policies, either expressed or implied, of the Department of the Air Force or the U.S. Government. The U.S. Government is authorized to reproduce and distribute reprints for Government purposes notwithstanding any copyright notation herein. The views and conclusions contained in this document are those of the authors
and should not be interpreted as representing the official
policies, either expressed or implied, of the Army Research
Office, Naval Research Office, Air Force, or the U.S. Government.
JH was supported by the Swiss National Science Foundation under NCCR Automation, grant agreement 51NF40~180545.
\looseness=-1

\bibliography{example_paper}

@article{chu2025sft,
  title={Sft memorizes, rl generalizes: A comparative study of foundation model post-training},
  author={Chu, Tianzhe and Zhai, Yuexiang and Yang, Jihan and Tong, Shengbang and Xie, Saining and Schuurmans, Dale and Le, Quoc V and Levine, Sergey and Ma, Yi},
  journal={arXiv preprint arXiv:2501.17161},
  year={2025}
}

@article{shenfeld2025rl,
  title={RL's Razor: Why Online Reinforcement Learning Forgets Less},
  author={Shenfeld, Idan and Pari, Jyothish and Agrawal, Pulkit},
  journal={arXiv preprint arXiv:2509.04259},
  year={2025}
}

@article{chen2025retaining,
  title={Retaining by Doing: The Role of On-Policy Data in Mitigating Forgetting},
  author={Chen, Howard and Razin, Noam and Narasimhan, Karthik and Chen, Danqi},
  journal={arXiv preprint arXiv:2510.18874},
  year={2025}
}

@article{huan2025does,
  title={Does Math Reasoning Improve General LLM Capabilities? Understanding Transferability of LLM Reasoning},
  author={Huan, Maggie and Li, Yuetai and Zheng, Tuney and Xu, Xiaoyu and Kim, Seungone and Du, Minxin and Poovendran, Radha and Neubig, Graham and Yue, Xiang},
  journal={arXiv preprint arXiv:2507.00432},
  year={2025}
}

@article{stiennon2020learning,
  title={Learning to summarize with human feedback},
  author={Stiennon, Nisan and Ouyang, Long and Wu, Jeffrey and Ziegler, Daniel and Lowe, Ryan and Voss, Chelsea and Radford, Alec and Amodei, Dario and Christiano, Paul F},
  journal={Advances in neural information processing systems},
  volume={33},
  pages={3008--3021},
  year={2020}
}

@inproceedings{ng2000algorithms,
  title={Algorithms for inverse reinforcement learning.},
  author={Ng, Andrew Y and Russell, Stuart and others},
  booktitle={Icml},
  volume={1},
  pages={2},
  year={2000}
}

@inproceedings{abbeel2004apprenticeship,
  title={Apprenticeship learning via inverse reinforcement learning},
  author={Abbeel, Pieter and Ng, Andrew Y},
  booktitle={Proceedings of the twenty-first international conference on Machine learning},
  pages={1},
  year={2004}
}

@article{peng2018deepmimic,
  title={Deepmimic: Example-guided deep reinforcement learning of physics-based character skills},
  author={Peng, Xue Bin and Abbeel, Pieter and Levine, Sergey and Van de Panne, Michiel},
  journal={ACM Transactions On Graphics (TOG)},
  volume={37},
  number={4},
  pages={1--14},
  year={2018},
  publisher={ACM New York, NY, USA}
}

@article{rafailov2023direct,
  title={Direct preference optimization: Your language model is secretly a reward model},
  author={Rafailov, Rafael and Sharma, Archit and Mitchell, Eric and Manning, Christopher D and Ermon, Stefano and Finn, Chelsea},
  journal={Advances in neural information processing systems},
  volume={36},
  pages={53728--53741},
  year={2023}
}

@inproceedings{korbak2022rl,
  title={RL with KL penalties is better viewed as Bayesian inference},
  author={Korbak, Tomasz and Perez, Ethan and Buckley, Christopher},
  booktitle={Findings of the Association for Computational Linguistics: EMNLP 2022},
  pages={1083--1091},
  year={2022}
}

@inproceedings{schulman2015trust,
  title={Trust region policy optimization},
  author={Schulman, John and Levine, Sergey and Abbeel, Pieter and Jordan, Michael and Moritz, Philipp},
  booktitle={International conference on machine learning},
  pages={1889--1897},
  year={2015},
  organization={PMLR}
}

@article{brown2020language,
  title={Language models are few-shot learners},
  author={Brown, Tom and Mann, Benjamin and Ryder, Nick and Subbiah, Melanie and Kaplan, Jared D and Dhariwal, Prafulla and Neelakantan, Arvind and Shyam, Pranav and Sastry, Girish and Askell, Amanda and others},
  journal={Advances in neural information processing systems},
  volume={33},
  pages={1877--1901},
  year={2020}
}

@article{snell2022learning,
  title={Learning by distilling context},
  author={Snell, Charlie and Klein, Dan and Zhong, Ruiqi},
  journal={arXiv preprint arXiv:2209.15189},
  year={2022}
}

@article{bai2022constitutional,
  title={Constitutional ai: Harmlessness from ai feedback},
  author={Bai, Yuntao and Kadavath, Saurav and Kundu, Sandipan and Askell, Amanda and Kernion, Jackson and Jones, Andy and Chen, Anna and Goldie, Anna and Mirhoseini, Azalia and McKinnon, Cameron and others},
  journal={arXiv preprint arXiv:2212.08073},
  year={2022}
}

@inproceedings{ross2011reduction,
  title={A reduction of imitation learning and structured prediction to no-regret online learning},
  author={Ross, St{\'e}phane and Gordon, Geoffrey and Bagnell, Drew},
  booktitle={Proceedings of the fourteenth international conference on artificial intelligence and statistics},
  pages={627--635},
  year={2011},
  organization={JMLR Workshop and Conference Proceedings}
}

@inproceedings{agarwal2024policy,
  title={On-policy distillation of language models: Learning from self-generated mistakes},
  author={Agarwal, Rishabh and Vieillard, Nino and Zhou, Yongchao and Stanczyk, Piotr and Garea, Sabela Ramos and Geist, Matthieu and Bachem, Olivier},
  booktitle={The Twelfth International Conference on Learning Representations},
  year={2024}
}

@article{tang2025few,
  title={On a few pitfalls in KL divergence gradient estimation for RL},
  author={Tang, Yunhao and Munos, R{\'e}mi},
  journal={arXiv preprint arXiv:2506.09477},
  year={2025}
}

@article{han2025general,
  title={General reasoning requires learning to reason from the get-go},
  author={Han, Seungwook and Pari, Jyothish and Gershman, Samuel J and Agrawal, Pulkit},
  journal={arXiv preprint arXiv:2502.19402},
  year={2025}
}

@article{li2025unveiling,
  title={Unveiling the Compositional Ability Gap in Vision-Language Reasoning Model},
  author={Li, Tianle and Zhang, Jihai and Rao, Yongming and Cheng, Yu},
  journal={arXiv preprint arXiv:2505.19406},
  year={2025}
}

@article{lai2025reinforcement,
  title={Reinforcement Fine-Tuning Naturally Mitigates Forgetting in Continual Post-Training},
  author={Lai, Song and Zhao, Haohan and Feng, Rong and Ma, Changyi and Liu, Wenzhuo and Zhao, Hongbo and Lin, Xi and Yi, Dong and Xie, Min and Zhang, Qingfu and others},
  journal={arXiv preprint arXiv:2507.05386},
  year={2025}
}

@article{hui2024qwen2,
  title={Qwen2. 5-coder technical report},
  author={Hui, Binyuan and Yang, Jian and Cui, Zeyu and Yang, Jiaxi and Liu, Dayiheng and Zhang, Lei and Liu, Tianyu and Zhang, Jiajun and Yu, Bowen and Lu, Keming and others},
  journal={arXiv preprint arXiv:2409.12186},
  year={2024}
}

@article{tang2023toolalpaca,
  title={Toolalpaca: Generalized tool learning for language models with 3000 simulated cases},
  author={Tang, Qiaoyu and Deng, Ziliang and Lin, Hongyu and Han, Xianpei and Liang, Qiao and Cao, Boxi and Sun, Le},
  journal={arXiv preprint arXiv:2306.05301},
  year={2023}
}

@article{feng2024sciknoweval,
  title={Sciknoweval: Evaluating multi-level scientific knowledge of large language models},
  author={Feng, Kehua and Ding, Keyan and Wang, Weijie and Zhuang, Xiang and Wang, Zeyuan and Qin, Ming and Zhao, Yu and Yao, Jianhua and Zhang, Qiang and Chen, Huajun},
  journal={arXiv preprint arXiv:2406.09098},
  year={2024}
}

@article{wu2025generalization,
  title={On the generalization of sft: A reinforcement learning perspective with reward rectification},
  author={Wu, Yongliang and Zhou, Yizhou and Ziheng, Zhou and Peng, Yingzhe and Ye, Xinyu and Hu, Xinting and Zhu, Wenbo and Qi, Lu and Yang, Ming-Hsuan and Yang, Xu},
  journal={arXiv preprint arXiv:2508.05629},
  year={2025}
}

@article{zellers2019hellaswag,
  title={Hellaswag: Can a machine really finish your sentence?},
  author={Zellers, Rowan and Holtzman, Ari and Bisk, Yonatan and Farhadi, Ali and Choi, Yejin},
  journal={arXiv preprint arXiv:1905.07830},
  year={2019}
}

@article{lin2021truthfulqa,
  title={Truthfulqa: Measuring how models mimic human falsehoods},
  author={Lin, Stephanie and Hilton, Jacob and Evans, Owain},
  journal={arXiv preprint arXiv:2109.07958},
  year={2021}
}

@article{hendrycks2020measuring,
  title={Measuring massive multitask language understanding},
  author={Hendrycks, Dan and Burns, Collin and Basart, Steven and Zou, Andy and Mazeika, Mantas and Song, Dawn and Steinhardt, Jacob},
  journal={arXiv preprint arXiv:2009.03300},
  year={2020}
}

@article{zhou2023instruction,
  title={Instruction-following evaluation for large language models},
  author={Zhou, Jeffrey and Lu, Tianjian and Mishra, Swaroop and Brahma, Siddhartha and Basu, Sujoy and Luan, Yi and Zhou, Denny and Hou, Le},
  journal={arXiv preprint arXiv:2311.07911},
  year={2023}
}

@article{sakaguchi2021winogrande,
  title={Winogrande: An adversarial winograd schema challenge at scale},
  author={Sakaguchi, Keisuke and Bras, Ronan Le and Bhagavatula, Chandra and Choi, Yejin},
  journal={Communications of the ACM},
  volume={64},
  number={9},
  pages={99--106},
  year={2021},
  publisher={ACM New York, NY, USA}
}

@article{chen2021evaluating,
  title={Evaluating large language models trained on code},
  author={Chen, Mark and Tworek, Jerry and Jun, Heewoo and Yuan, Qiming and Pinto, Henrique Ponde De Oliveira and Kaplan, Jared and Edwards, Harri and Burda, Yuri and Joseph, Nicholas and Brockman, Greg and others},
  journal={arXiv preprint arXiv:2107.03374},
  year={2021}
}

@article{lu2025onpolicydistillation,
  author = {Kevin Lu and Thinking Machines Lab},
  title = {On-Policy Distillation},
  journal = {Thinking Machines Lab: Connectionism},
  year = {2025},
  note = {https://thinkingmachines.ai/blog/on-policy-distillation},
  doi = {10.64434/tml.20251026},
}

@article{yue2025does,
  title={Does reinforcement learning really incentivize reasoning capacity in llms beyond the base model?},
  author={Yue, Yang and Chen, Zhiqi and Lu, Rui and Zhao, Andrew and Wang, Zhaokai and Song, Shiji and Huang, Gao},
  journal={arXiv preprint arXiv:2504.13837},
  year={2025}
}

@article{wu2025invisible,
  title={The invisible leash: Why rlvr may or may not escape its origin},
  author={Wu, Fang and Xuan, Weihao and Lu, Ximing and Liu, Mingjie and Dong, Yi and Harchaoui, Zaid and Choi, Yejin},
  journal={arXiv preprint arXiv:2507.14843},
  year={2025}
}

@article{mecklenburg2024injecting,
  title={Injecting new knowledge into large language models via supervised fine-tuning},
  author={Mecklenburg, Nick and Lin, Yiyou and Li, Xiaoxiao and Holstein, Daniel and Nunes, Leonardo and Malvar, Sara and Silva, Bruno and Chandra, Ranveer and Aski, Vijay and Yannam, Pavan Kumar Reddy and others},
  journal={arXiv preprint arXiv:2404.00213},
  year={2024}
}

@article{kujanpaa2025efficient,
  title={Efficient knowledge injection in LLMs via self-distillation},
  author={Kujanp{\"a}{\"a}, Kalle and Marttinen, Pekka and Valpola, Harri and Ilin, Alexander},
  journal={Transactions on Machine Learning Research},
  year={2025}
}

@article{eyuboglu2025cartridges,
  title={Cartridges: Lightweight and general-purpose long context representations via self-study},
  author={Eyuboglu, Sabri and Ehrlich, Ryan and Arora, Simran and Guha, Neel and Zinsley, Dylan and Liu, Emily and Tennien, Will and Rudra, Atri and Zou, James and Mirhoseini, Azalia and others},
  journal={arXiv preprint arXiv:2506.06266},
  year={2025}
}

@article{lazzati2024does,
  title={How does inverse rl scale to large state spaces? a provably efficient approach},
  author={Lazzati, Filippo and Mutti, Mirco and Metelli, Alberto Maria},
  journal={Advances in Neural Information Processing Systems},
  volume={37},
  pages={54820--54871},
  year={2024}
}

@article{arora2021survey,
  title={A survey of inverse reinforcement learning: Challenges, methods and progress},
  author={Arora, Saurabh and Doshi, Prashant},
  journal={Artificial Intelligence},
  volume={297},
  pages={103500},
  year={2021},
  publisher={Elsevier}
}

@inproceedings{ziebart2008maximum,
  title={Maximum entropy inverse reinforcement learning.},
  author={Ziebart, Brian D and Maas, Andrew L and Bagnell, J Andrew and Dey, Anind K and others},
  booktitle={Aaai},
  volume={8},
  pages={1433--1438},
  year={2008},
  organization={Chicago, IL, USA}
}

@article{wulfmeier2015maximum,
  title={Maximum entropy deep inverse reinforcement learning},
  author={Wulfmeier, Markus and Ondruska, Peter and Posner, Ingmar},
  journal={arXiv preprint arXiv:1507.04888},
  year={2015}
}

@article{ho2016generative,
  title={Generative adversarial imitation learning},
  author={Ho, Jonathan and Ermon, Stefano},
  journal={Advances in neural information processing systems},
  volume={29},
  year={2016}
}

@article{ouyang2022training,
  title={Training language models to follow instructions with human feedback},
  author={Ouyang, Long and Wu, Jeffrey and Jiang, Xu and Almeida, Diogo and Wainwright, Carroll and Mishkin, Pamela and Zhang, Chong and Agarwal, Sandhini and Slama, Katarina and Ray, Alex and others},
  journal={Advances in neural information processing systems},
  volume={35},
  pages={27730--27744},
  year={2022}
}

@article{ziegler2019fine,
  title={Fine-tuning language models from human preferences},
  author={Ziegler, Daniel M and Stiennon, Nisan and Wu, Jeffrey and Brown, Tom B and Radford, Alec and Amodei, Dario and Christiano, Paul and Irving, Geoffrey},
  journal={arXiv preprint arXiv:1909.08593},
  year={2019}
}

@article{xu2020error,
  title={Error bounds of imitating policies and environments},
  author={Xu, Tian and Li, Ziniu and Yu, Yang},
  journal={Advances in Neural Information Processing Systems},
  volume={33},
  pages={15737--15749},
  year={2020}
}

@article{chen2024huatuogpt,
  title={Huatuogpt-o1, towards medical complex reasoning with llms},
  author={Chen, Junying and Cai, Zhenyang and Ji, Ke and Wang, Xidong and Liu, Wanlong and Wang, Rongsheng and Hou, Jianye and Wang, Benyou},
  journal={arXiv preprint arXiv:2412.18925},
  year={2024}
}

@inproceedings{amini2025better,
  title={Better Estimation of the Kullback--Leibler Divergence Between Language Models},
  author={Amini, Afra and Vieira, Tim and Cotterell, Ryan},
  booktitle={The Thirty-ninth Annual Conference on Neural Information Processing Systems},
  year={2025}
}

@article{mitra2025semantic,
  title={Semantic Soft Bootstrapping: Long Context Reasoning in LLMs without Reinforcement Learning},
  author={Mitra, Purbesh and Ulukus, Sennur},
  journal={arXiv preprint arXiv:2512.05105},
  year={2025}
}

@book{sutton1998reinforcement,
  title={Reinforcement learning: An introduction},
  author={Sutton, Richard S and Barto, Andrew G and others},
  volume={1},
  year={1998},
  publisher={MIT press Cambridge}
}

@article{olmo2025olmo,
  title={Olmo 3},
  author={Olmo, Team and Ettinger, Allyson and Bertsch, Amanda and Kuehl, Bailey and Graham, David and Heineman, David and Groeneveld, Dirk and Brahman, Faeze and Timbers, Finbarr and Ivison, Hamish and others},
  journal={arXiv preprint arXiv:2512.13961},
  year={2025}
}

@article{hassabis2017neuroscience,
  title={Neuroscience-inspired artificial intelligence},
  author={Hassabis, Demis and Kumaran, Dharshan and Summerfield, Christopher and Botvinick, Matthew},
  journal={Neuron},
  volume={95},
  number={2},
  pages={245--258},
  year={2017},
  publisher={Elsevier}
}

@article{de2021continual,
  title={A continual learning survey: Defying forgetting in classification tasks},
  author={De Lange, Matthias and Aljundi, Rahaf and Masana, Marc and Parisot, Sarah and Jia, Xu and Leonardis, Ale{\v{s}} and Slabaugh, Gregory and Tuytelaars, Tinne},
  journal={IEEE transactions on pattern analysis and machine intelligence},
  volume={44},
  number={7},
  pages={3366--3385},
  year={2021},
  publisher={IEEE}
}

@misc{eval-harness,
  author       = {Gao, Leo and Tow, Jonathan and Abbasi, Baber and Biderman, Stella and Black, Sid and DiPofi, Anthony and Foster, Charles and Golding, Laurence and Hsu, Jeffrey and Le Noac'h, Alain and Li, Haonan and McDonell, Kyle and Muennighoff, Niklas and Ociepa, Chris and Phang, Jason and Reynolds, Laria and Schoelkopf, Hailey and Skowron, Aviya and Sutawika, Lintang and Tang, Eric and Thite, Anish and Wang, Ben and Wang, Kevin and Zou, Andy},
  title        = {The Language Model Evaluation Harness},
  month        = 07,
  year         = 2024,
  publisher    = {Zenodo},
  version      = {v0.4.3},
  doi          = {10.5281/zenodo.12608602},
  url          = {https://zenodo.org/records/12608602}
}

@article{hinton2015distilling,
  title={Distilling the knowledge in a neural network},
  author={Hinton, Geoffrey and Vinyals, Oriol and Dean, Jeff},
  journal={arXiv preprint arXiv:1503.02531},
  year={2015}
}

@article{kirkpatrick2017overcoming,
  title={Overcoming catastrophic forgetting in neural networks},
  author={Kirkpatrick, James and Pascanu, Razvan and Rabinowitz, Neil and Veness, Joel and Desjardins, Guillaume and Rusu, Andrei A and Milan, Kieran and Quan, John and Ramalho, Tiago and Grabska-Barwinska, Agnieszka and others},
  journal={Proceedings of the national academy of sciences},
  volume={114},
  number={13},
  pages={3521--3526},
  year={2017},
  publisher={National Academy of Sciences}
}

@article{li2017learning,
  title={Learning without forgetting},
  author={Li, Zhizhong and Hoiem, Derek},
  journal={IEEE transactions on pattern analysis and machine intelligence},
  volume={40},
  number={12},
  pages={2935--2947},
  year={2017},
  publisher={IEEE}
}

@article{yang2024self,
  title={Self-distillation bridges distribution gap in language model fine-tuning},
  author={Yang, Zhaorui and Pang, Tianyu and Feng, Haozhe and Wang, Han and Chen, Wei and Zhu, Minfeng and Liu, Qian},
  journal={arXiv preprint arXiv:2402.13669},
  year={2024}
}

@article{swamy2022sequence,
  title={Sequence model imitation learning with unobserved contexts},
  author={Swamy, Gokul and Choudhury, Sanjiban and Bagnell, J and Wu, Steven Z},
  journal={Advances in Neural Information Processing Systems},
  volume={35},
  pages={17665--17676},
  year={2022}
}

@article{padmanabhan2023propagating,
  title={Propagating knowledge updates to lms through distillation},
  author={Padmanabhan, Shankar and Onoe, Yasumasa and Zhang, Michael and Durrett, Greg and Choi, Eunsol},
  journal={Advances in Neural Information Processing Systems},
  volume={36},
  pages={47124--47142},
  year={2023}
}

@inproceedings{cao2025infiniteicl,
  title={Infiniteicl: Breaking the limit of context window size via long short-term memory transformation},
  author={Cao, Bowen and Cai, Deng and Lam, Wai},
  booktitle={Findings of the Association for Computational Linguistics: ACL 2025},
  pages={11402--11415},
  year={2025}
}

@article{zhao2026self,
  title={Self-distilled reasoner: On-policy self-distillation for large language models},
  author={Zhao, Siyan and Xie, Zhihui and Liu, Mengchen and Huang, Jing and Pang, Guan and Chen, Feiyu and Grover, Aditya},
  journal={arXiv preprint arXiv:2601.18734},
  year={2026}
}

@article{hubotter2026reinforcement,
  title={Reinforcement learning via self-distillation},
  author={H{\"u}botter, Jonas and L{\"u}beck, Frederike and Behric, Lejs and Baumann, Anton and Bagatella, Marco and Marta, Daniel and Hakimi, Ido and Shenfeld, Idan and Buening, Thomas Kleine and Guestrin, Carlos and others},
  journal={arXiv preprint arXiv:2601.20802},
  year={2026}
}
\bibliographystyle{iclr2025_conference}

\newpage
\appendix

\section{Additional Ablations}

\subsection{Estimating the KL gradient}
\label{app:gradients}
A central component of our objective is the gradient of the KL divergence between the current policy $\pi_\theta(y|x)$ and the teacher policy $\pi(y|x,c)$. For sequence models, the KL divergence is defined at the sequence level as:
$$
\mathrm{KL}(\pi_\theta ||\pi)=\mathbb{E}_{y \sim \pi_\theta}
\left[
\log \frac{\pi_\theta(y|x)}{\pi(y|x,c)}
\right]
$$
where $y = (y_1,\dots,y_T)$ is generated autoregressively. Differentiating this quantity is non-trivial because $\pi_\theta$ appears both in the sampling distribution and inside the logarithm, and different practical estimators trade off bias, variance, and computational cost \citep{tang2025few}.

We consider and ablate several commonly used KL gradient estimators.

\paragraph{Token-level (partial) estimator}. A widely used approximation decomposes the KL into token-level terms and differentiates each independently:
$$
\widehat{g}_{\text{token}}= \sum_{t=1}^T
\log \frac{\pi_\theta(y_t|y_{<t},x)}{\pi(y_t|y_{<t},x,c)}
\nabla_\theta \log\pi_\theta(y_t|y_{<t},x)
$$
As shown in recent analyses, this estimator corresponds to a \textit{partial} derivative of the sequence-level KL: it ignores the effect of early tokens on future token distributions and is therefore biased with respect to the true gradient \citep{tang2025few}.

\paragraph{Full analytic per-token estimator.} An alternative is to compute the KL analytically at each timestep by marginalizing over the vocabulary:
$$
\widehat{g}_{\text{analytic}}=
\sum_{t=1}^T
\sum_{v \in \mathcal{V}}
\log \frac{\pi_\theta(v|y_{<t},x)}{\pi(v|y_{<t},x,c)}
\nabla_\theta \log\pi_\theta(v|y_{<t},x)
\label{eq:g_analytic}
$$
This estimator has strictly lower variance than sample-based token estimators, but it remains biased at the \textit{sequence} level, since it still does not account for how the choice of $y_t$ influences future states $y_{>t}$. Despite this bias, it is often computationally attractive because it leverages quantities already produced during the forward pass.

\paragraph{Rao-Blackwellized estimator.} Following recent work \citep{amini2025better}, one can further reduce variance by Rao-Blackwellizing the KL estimator, analytically integrating over next-token distributions while retaining Monte-Carlo sampling over prefixes. This yields an unbiased estimator of both the KL and its gradient with provably lower variance than standard Monte-Carlo estimators . However, this estimator is more expensive to compute.

$$
\widehat{g}_{\text{rb}}=
\sum_{t=1}^{T}\left[\sum_{v\in\mathcal V}\log \frac{\pi_\theta(v|y_{<t},x)}{\pi(v|y_{<t},x,c)}
\nabla_\theta \log\pi_\theta(v|y_{<t},x)+k_\theta(y_{<t})\sum_{i=1}^{t-1}\nabla_\theta \log \pi_\theta(y_i|y_{<i},x)\right]
$$

Where $k_\theta(y_{<t})$ is the stepwise KL term $\mathrm{KL}(\pi_\theta(\cdot|y_{<t},x) ||\pi(\cdot|y_{<t},x,x))$.

We empirically ablate all three estimators in our training pipeline. Despite its theoretical bias, we find that the \textit{full analytic per-token estimator} consistently yields the most stable optimization and best downstream performance. In contrast, the token-level estimator exhibits higher variance and weaker KL control, while the Rao–Blackwellized estimator did not provide measurable gains in our setting relative to its additional complexity.

We also experimented with drawing multiple trajectories per prompt to reduce variance in the gradient estimator, which is theoretically beneficial for Monte-Carlo estimates. In practice, however, increasing the number of samples per prompt produced negligible improvements while substantially increasing compute. As a result, we adopt a single-trajectory-per-prompt setup combined with the analytic per-token KL estimator in all main experiments.

\subsection{The Importance of Demonstration-Conditioned Context}

We analyze which components of the teacher context are essential for the effectiveness of our method in the \emph{Knowledge Acquisition} setting. Recent self-distillation approaches for knowledge injection perform \emph{offline} distillation using only the raw corpus as context \citep{eyuboglu2025cartridges, kujanpaa2025efficient}. In contrast, our approach differs along two dimensions: (i) the teacher is conditioned not only on the source text but also on a worked answer, and (ii) distillation is performed on-policy.

\begin{figure}[h]
    \centering
    \includegraphics[width=0.48\textwidth]{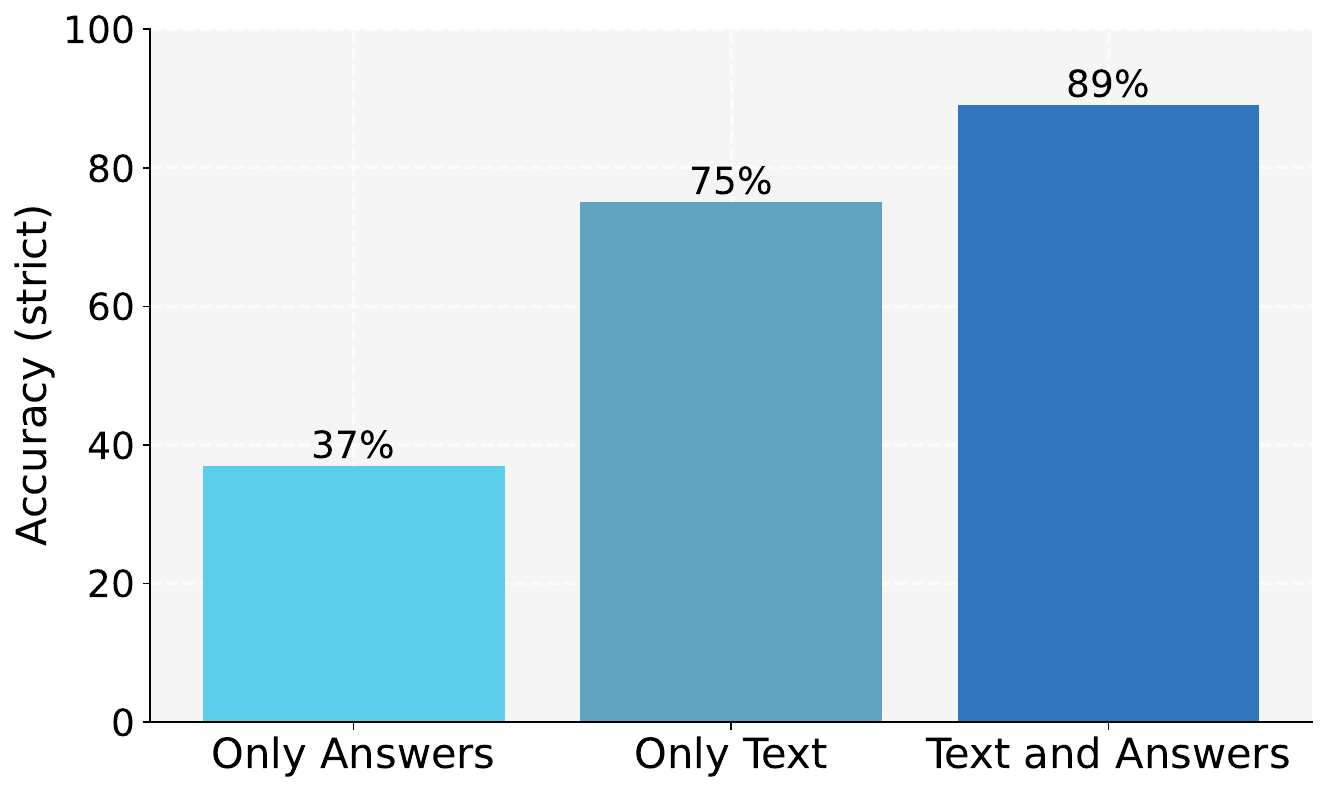}
    \caption{Conditioning the teacher on both article text and answer (89\% strict accuracy) substantially outperforms text-only conditioning (75\%), showing that the full demonstration context is critical for effective knowledge transfer.}
    \label{fig:KI_ablation}
\end{figure}

In this subsection, we isolate the effect of the teacher context while holding the on-policy training procedure fixed. Specifically, we compare three variants: conditioning the teacher on \emph{only the article text}, \emph{only the answer}, and the full \emph{text-plus-answer} context. A direct comparison to offline distillation methods is deferred to Section~\ref{subsec:ablation}.

\paragraph{Results.}
The results are shown in Figure~\ref{fig:KI_ablation}. Conditioning the teacher on the full text-plus-answer context yields the strongest performance, achieving 89\% strict accuracy. Using only the article text substantially underperforms, consistent with prior findings that text-only distillation provides a weak and noisy supervisory signal. Conditioning on text alone improves performance over answers-only context but still falls short of the full context.
These results suggest that answer-conditioned context plays a critical role by providing stronger guidance for the student policy.

\subsection{Choice of teacher model}
\label{app:ema_ablation}
We ablate the choice of teacher policy used for distillation. While our framework does not require an external teacher, the stability of training depends critically on how the teacher is instantiated.

\begin{figure}[h]
    \centering
    \includegraphics[width=0.48\textwidth]{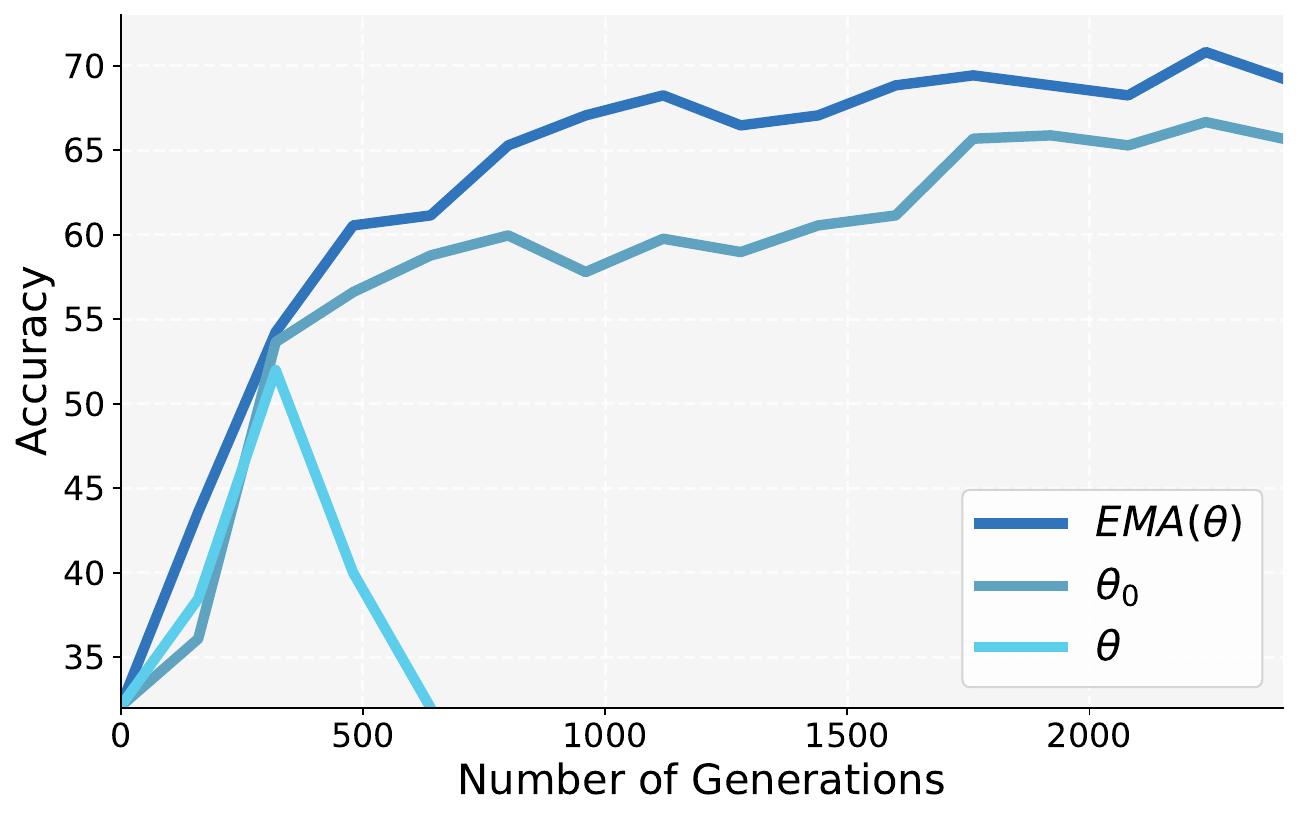}
    \caption{EMA teacher provides stable and effective training.  Using the frozen base model, which fails to track learning progress lead to inferior results. Using the current student directly as the teacher leads to training instabilities.}
    \label{fig:ema_ablation}
\end{figure}

Using the frozen base model as the teacher yields stable training but consistently underperforms, as the teacher fails to reflect improvements acquired during learning. At the other extreme, using the student model itself as the teacher leads to severe instabilities. In this setting, small stochastic fluctuations in token-level probability updates can be rapidly amplified through the on-policy feedback loop, causing training to diverge.
We find that maintaining an exponential moving average (EMA) of the student parameters provides an effective compromise. As shown in Figure~\ref{fig:ema_ablation}, the EMA teacher tracks the student’s progress while smoothing high-variance updates, resulting in both stable training and superior final performance. 

\section{Training and Evaluation details}
\label{app:hps}
\subsection{Training Details}
All experiments were conducted using the Hugging Face TRL library. Each experiment was conducted on a single NVIDIA H200 GPU. We perform full fine-tuning of the entire model's parameters. For each method, we performed a hyperparameter sweep over learning rates, batch sizes, and training epochs. We report test results for the model checkpoint that achieved the best validation performance on the target task.

Tables \ref{tab:hyperparameters_skill} and \ref{tab:hyperparameters_knowledge} present the full hyperparameter search spaces and final selected values for the Skill Learning and Knowledge Acquisition settings, respectively. Across all tasks, we found that SDFT benefits from training for multiple epochs—typically 2 epochs for Skill Learning tasks and 4 epochs for Knowledge Acquisition. In contrast, SFT tends to overfit rapidly and showed no performance gains beyond a single epoch in most cases.

For SDFT, The teacher context was constructed using the prompt template shown in Section \ref{subsec:algo}. We employed the analytic per-token KL gradient estimator (see Appendix \ref{app:gradients}) with a single on-policy rollout per training example.

\subsection{Evaluation Details}
\textbf{Sampling Strategy.} For accuracy metrics, we used greedy decoding (temperature = 0). For pass@k experiments, we used temperature = 1.0 with nucleus sampling (top-p = 0.95).

\textbf{Statistical Reporting.} Unless mentioned otherwise, all experiments were run over 3 random seeds. We report mean performance and 95\% confidence intervals across seeds.

\textbf{Prior Capabilities Evaluation.} We assessed performance on general capabilities using the suite of benchmarks described in Section \ref{subsec:exp_setting}: HellaSwag \citep{zellers2019hellaswag}, TruthfulQA \citep{lin2021truthfulqa}, MMLU \citep{hendrycks2020measuring}, IFEval \citep{zhou2023instruction}, Winogrande \citep{sakaguchi2021winogrande}, and HumanEval \citep{chen2021evaluating}. All benchmark evaluations were conducted using the Language Model Evaluation Harness \citep{eval-harness}.

\subsection{Dataset Details}
\paragraph{Science Q\&A.} We used the Chemistry L-3 subset from SciKnowEval \citep{feng2024sciknoweval}, splitting the data into approximately 75\% train, 5\% validation, and 20\% test. To construct expert demonstrations, we queried GPT-4o, sampling up to 8 responses per prompt and retaining a single response that matched the correct final answer. This procedure yielded valid demonstrations for 100\% of training examples. Since this is a multiple-choice dataset, accuracy was computed by exact match between the model's final answer choice and the ground truth.
\paragraph{Tool Use.} We used the ToolAlpaca dataset \citep{tang2023toolalpaca}, following the original train-test split provided by the authors. Expert demonstrations were included in the original dataset. Accuracy was evaluated using regex matching against the ground-truth API call, accounting for variations in argument ordering.
\paragraph{Medical.} We built upon the HuatuoGPT-o1 dataset \citep{chen2024huatuogpt}, which provides both an SFT training set and a collection of problems with only final answer (used in the original paper for RL training). For training, we used only the English-language questions, yielding approximately 20,000 examples. For evaluation, we randomly sampled 1,000 verifiable questions from the verifiable problem set. Since these are open-ended clinical reasoning questions, we used GPT-5-mini as an automated evaluator with the following prompt:
\begin{tcolorbox}
\begin{verbatim}
You are an expert medical evaluator assessing whether a model's 
response correctly answers a medical question. Your task is to 
compare the model's response to the reference answer and determine
if the model's response is:
1. CORRECT: The response contains the key medical information from
the  reference answer, even if phrased differently or includes 
additional correct medical details.
2. INCORRECT: The response is medically wrong, misses the main 
point, or provides incorrect medical information.
Focus on medical accuracy and completeness, not on writing style or
verbosity.

[Medical Question]
{question}
[Reference Answer]
{reference_answer}
[Model Response]
{model_response}
Evaluate the model's response. Output ONLY one of: "CORRECT" or
"INCORRECT".
\end{verbatim}
\end{tcolorbox}

\paragraph{Knowledge Acquisition.} We constructed a corpus of Wikipedia articles describing natural disasters that occurred in 2025 (after the model's knowledge cutoff), including:

2025 Myanmar earthquake, 2025 Kamchatka earthquake, 2025 Uttarakhand flash flood, Typhoon Kalmaegi, Tropical Storm Wipha, Cyclonic Ditwah, Hurricane Melissa, Kentwood Carson Tornado, July 2025 Central Texas floods.

Following \citet{mecklenburg2024injecting}, we used GPT-5 to generate question-answer pairs from these articles, using the following prompt:
\begin{tcolorbox}
\begin{verbatim}
You are a helpful assistant that helps me write questions for an
exam. You  will be given a wiki article and you will need to write 
100 question on the content of the wiki article. The question should
require recalling multiple pieces of information from the wiki
article. Do not repeat the same question

The questions should be in the following format:
Question: <question>
Answer: <answern>
\end{verbatim}
\end{tcolorbox}

We also manually verify that the same question wasn't generated more than once. For evaluation, we used GPT-5-mini as an automated evaluator with the following prompt:

\begin{tcolorbox}
\begin{verbatim}
You are an expert evaluator assessing whether a model's response 
correctly answers a question.

Your task is to compare the model's response to the reference answer
and determine if the model's response is:
1. CORRECT: The response contains the key information from the 
reference answer, even if phrased differently or includes additional
correct details.
2. PARTIALLY_CORRECT: The response contains most of the key 
information from the reference answer but misses some details.
3. INCORRECT: The response is wrong, misses the main point, or 
provides incorrect information.

Focus on factual accuracy and completeness, not on writing style or
verbosity.

[Question]
{question}

[Reference Answer]
{reference_answer}

[Model Response]
{model_response}

Evaluate the model's response. Output ONLY one of: 
"CORRECT", "PARTIALLY_CORRECT", or "INCORRECT".
\end{verbatim}
\end{tcolorbox}

\begin{algorithm}[t]
\caption{Self-Distillation Fine-Tuning (SDFT)}
\label{alg:sdft}
\begin{algorithmic}[1]
\REQUIRE Demonstration dataset $\mathcal{D}=\{(x_i,c_i)\}_{i=1}^N$
\REQUIRE Autoregressive model $\pi_\theta$; student context prompt $\mathrm{Ctx}_S(x)$; teacher context prompt $\mathrm{Ctx}_T(x,c)$
\REQUIRE Batch size $B$, max generation length $T$, learning rate $\eta$, teacher EMA rate $\alpha$
\STATE Set teacher weights $\phi=\theta$.
\FOR{each training step}
  \STATE Sample minibatch $\mathcal{B}=\{(x_i,c_i)\}_{i=1}^B \sim \mathcal{D}$
  \FORALL{$(x_i,c_i)\in \mathcal{B}$ \textbf{in parallel}}
    \STATE \textbf{Student rollout (on-policy):}
    \STATE $s_i \gets \mathrm{Ctx}_S(x_i)$
    \STATE Sample $y_i=(y_{i,1:T}) \sim P_{\text{sample}}(\cdot \mid s_i)$
    \STATE \textbf{Compute teacher and student token logprobs on the sampled tokens:}
    \STATE $t_i \gets \mathrm{Ctx}_T(x_i,c_i)$
    \STATE Using $\mathrm{TrainEngine}$, compute
    \STATE \hspace{1.5em} $\ell^{S}_{i,t} \gets \log \pi_{\theta}(y_{i,t}\mid y_{i,<t}, s_i)$ and
    \STATE \hspace{1.5em} $\ell^{T}_{i,t} \gets \log \pi_{\phi}(y_{i,t}\mid y_{i,<t}, t_i)$
  \ENDFOR
  \STATE \textbf{Gradient computation and update:}
  \STATE Compute gradient estimate using Eq.~\ref{eq:g_analytic}:
  \STATE \hspace{1.5em} $g \gets \frac{1}{B}\sum_{i=1}^B g_{\text{analytic}}\!\Big(\{(\ell^{S}_{i,t},\ell^{T}_{i,t})\}_{t=1}^{T}\Big)$
  \STATE If needed, add importance sampling to compensate for differences between the inference engine (e.g., VLLM) and the training code.
  \STATE Update parameters: $\theta \gets \theta - \eta \, g$
  \STATE Update teacher parameters: $\phi \gets\alpha \theta + (1-\alpha)\phi$
\ENDFOR
\end{algorithmic}
\end{algorithm}

\begin{table}[h]
\centering
\begin{tabular}{@{}llll@{}}
\toprule
\textbf{Hyperparameter}         & \textbf{SFT}                            & \textbf{DFT}       & \textbf{SDFT}                         \\ \midrule
Base Model             & Qwen2.5 7B-Instruct                    & Qwen2.5 7B-Instruct         & Qwen2.5 7B-Instruct            \\
Learning Rate          & \{5e-6, 1e-5, 5e-5\}    & \{5e-6, 1e-5, 5e-5\}& \{5e-6, 1e-5, 5e-5\} \\
Optimizer              & adamw           & adamw       & adamw          \\
LR Scheduler           & Cosine w. warmup & Cosine w. warmup            & Cosine w. warmup      \\
Warmup steps           & 10                                     & 10     & 10                   \\
Epochs                 & \{1,2\}                                       & \{1,2\}           & \{1,2\}          \\
Batch Size             & \{16,32,64\}        & \{16,32,64\}      & \{16,32,64\}    \\
Max Grad Norm          & 1                                      & 1                  & 1                \\
bfloat16               & True        & True              & True                           \\
Weight Decay               & 0        & 0       & 0                           \\
\multicolumn{3}{l}{\textit{SDFT-only hyperparameters}}                                                    \\
EMA $\alpha$                &                &      & \{0.01, 0.02, 0.05\}                              \\
Max generation length                &                &      & 2048                              \\  \bottomrule
\end{tabular}
\caption{Hyperparameters used for the Skill Learning experiments. Curly braces \{\} indicate a sweep over the specified values.}
\label{tab:hyperparameters_skill}
\end{table}

\begin{table}[h]
\centering
\begin{tabular}{@{}llll@{}}
\toprule
\textbf{Hyperparameter}         & \textbf{SFT}                            & \textbf{CPT}       & \textbf{SDFT}                         \\ \midrule
Base Model             & Qwen2.5 7B-Instruct                    & Qwen2.5 7B-Instruct         & Qwen2.5 7B-Instruct            \\
Learning Rate          & \{5e-6, 1e-5, 5e-5\}    & \{1e-6, 5e-6, 1e-5\}& \{5e-6, 1e-5, 5e-5\} \\
Optimizer              & adamw           & adamw       & adamw          \\
LR Scheduler           & Cosine w. warmup & Cosine w. warmup            & Cosine w. warmup      \\
Warmup steps           & 10                                     & 10     & 10                   \\
Epochs                 & \{1,2\}                                       & \{1,2,4,8\}           & \{1,2,4\}          \\
Batch Size             &  \{16,32,64\}    & N/A      & \{16,32,64\}    \\
Max Grad Norm          & 1                                      & 1                  & 1                \\
bfloat16               & True        & True              & True                           \\
Weight Decay               & 0        & 0       & 0                           \\
\multicolumn{3}{l}{\textit{SDFT-only hyperparameters}}                                                    \\
EMA $\alpha$                &                &      & \{0.01, 0.02, 0.05\}                              \\
Max generation length                &                &      & 1024                              \\  \bottomrule
\end{tabular}
\caption{Hyperparameters used for the Knowledge Acquisition experiments. Curly braces \{\} indicate a sweep over the specified values.}
\label{tab:hyperparameters_knowledge}
\end{table}

\begin{table*}[t]
\centering
\resizebox{\textwidth}{!}{
\begin{tabular}{@{}lcccccccc@{}}
\toprule
 & New Task:     & \multicolumn{7}{c}{Previous Tasks:}                                     \\ 
      & Science Q\&A & Hellaswag & Humaneval & IFeval & MMLU & TruthfulQA & Winogrande & Avg. \\ \midrule
Base (Qwen2.5-7B)  & \multicolumn{1}{|c|}{32.1}         & \textbf{62.0}      & 65.8      & \textbf{74.3}   & \textbf{71.7} & \textbf{47.9}       & 71.1       & \textbf{65.5} \\
SFT   & \multicolumn{1}{|c|}{\underline{66.2}}         &  55.0      &  54.8      &  35.3      &   64.6 &    36.8     &   \textbf{73.7}      & 53.4   \\
SFT + re-invoke  & \multicolumn{1}{|c|}{66.0}         &  \underline{61.6}      &  63.4      &  52.9      &   68.7 &    45.2     &   70.0      & 60.2   \\
DFT   & \multicolumn{1}{|c|}{54.8}         &  57.6      &  \underline{67.0}      &  60.4      &   69.4 &    38.8     &   68.2      & 60.2   \\
SDFT (Ours)  & \multicolumn{1}{|c|}{\textbf{70.2}}         & 60.9      & \textbf{68.9}      & 66.8   & \underline{70.7} & \underline{46.5}       & \underline{73.1}       & \underline{64.5} \\ \bottomrule
\end{tabular}
}
\vspace{5pt} \newline \vspace{5pt}
\centering
\resizebox{\textwidth}{!}{
\begin{tabular}{@{}lcccccccc@{}}
\toprule
 & New Task: & \multicolumn{7}{c}{Previous Tasks:}                                     \\ 
      & Tooluse  & Hellaswag & Humaneval & IFeval & MMLU & TruthfulQA & Winogrande & Avg. \\ \midrule
Base (Qwen2.5-7B)  
& \multicolumn{1}{|c|}{42.9}      & \textbf{62.0}      & 65.8      & \textbf{74.3}   & \textbf{71.7} & \underline{47.9}       & 71.1       & \textbf{65.5} \\
SFT   & \multicolumn{1}{|c|}{63.2}     & 57.3      & 50.0      & 49.8   & 70.2 & 37.5       & \textbf{73.1}       & 56.0 \\
SFT + re-invoke  & \multicolumn{1}{|c|}{63.1}         & \underline{61.7}      & \textbf{68.9}      & 59.1      & 71.5 & \textbf{49.1}    & 71.6      & 63.7   \\
DFT    & \multicolumn{1}{|c|}{\underline{64.2}}     & 59.7      & 61.4      & 60.2   & 71.6 & 40.2       & 71.5       & 60.8 \\
SDFT (Ours)  & \multicolumn{1}{|c|}{\textbf{70.6}}     & 61.6      & \underline{68.3}      & \underline{71.9}   & \underline{71.5} & 47.3       & \underline{71.7}       & \underline{65.4} \\ 
\bottomrule
\end{tabular}
}
\vspace{5pt} \newline \vspace{5pt}
\centering
\resizebox{\textwidth}{!}{
\begin{tabular}{@{}lcccccccc@{}}
\toprule
 & New Task: & \multicolumn{7}{c}{Previous Tasks:}                                     \\ 
      & Medical  & Hellaswag & Humaneval & IFeval & MMLU & TruthfulQA & Winogrande & Avg. \\ \midrule
Base (Qwen2.5-7B)  & \multicolumn{1}{|c|}{30.1}     & \textbf{62.0}      & \underline{65.8}      & \underline{74.3}   & \textbf{71.7} & \textbf{47.9}       & 71.1       & \textbf{65.5} \\
SFT   & \multicolumn{1}{|c|}{35.5}        &   59.5    &   62.1    &  56.6  & 70.5 &   39.8     &  \textbf{72.9}      & 60.2 \\
SFT + re-invoke  &   \multicolumn{1}{|c|}{35.6}     &   61.5    &   63.1    &  67.6  & 70.0 &   42.3     &  71.4      & 62.6 \\
DFT   &   \multicolumn{1}{|c|}{\underline{36.2}}     &  \underline{61.9}     &   64.6    &  \textbf{74.6}  & \underline{71.6}  &     40.1  &   71.3     & 64.0 \\
SDFT (Ours)  & \multicolumn{1}{|c|}{\textbf{40.2}}        &   61.4    &    67.7   & 72.3   & 71.5 &  \underline{47.3}      & \underline{71.9}       &  \underline{65.4} \\ \bottomrule
\end{tabular}
}
\caption{The table reports the exact new-task accuracy and average prior-task performance for each method across all Skill Learning tasks.}
\label{tab:skills}
\end{table*}

\end{document}